\documentclass{article} %
\usepackage[dvipsnames]{xcolor}

\usepackage{iclr2025_conference,times}

\usepackage{amsmath,amsfonts,bm}

\def\eqref#1{equation~\ref{#1}}

\def\1{\bm{1}}

\DeclareMathAlphabet{\mathsfit}{\encodingdefault}{\sfdefault}{m}{sl}
\SetMathAlphabet{\mathsfit}{bold}{\encodingdefault}{\sfdefault}{bx}{n}

\usepackage[breaklinks=true,bookmarks=false,colorlinks=true,pagebackref=true,citecolor=RoyalBlue]{hyperref}
\usepackage{url}

\usepackage[utf8]{inputenc} %
\usepackage[T1]{fontenc}    %
\usepackage{hyperref}       %
\usepackage{url}            %
\usepackage{booktabs}       %
\usepackage{amsfonts}       %
\usepackage{nicefrac}       %
\usepackage{microtype}      %
\usepackage{xcolor}         %
\usepackage{caption}
\usepackage{amsmath}
\usepackage{amssymb}
\usepackage{booktabs}
\usepackage{cuted}
\usepackage{multirow}
\usepackage{tabularx}
\usepackage{pifont}
\usepackage{wrapfig}
\usepackage{graphicx}

\definecolor{DarkGreen}{RGB}{21,100,52}
\definecolor{DarkRed}{RGB}{139,0,0}
\definecolor{DarkRed2}{RGB}{180,0,0}
\definecolor{MyGreen}{rgb}{0, 0.55, 0}

\newif\ifdrafting
\draftingtrue %
\ifdrafting
\newcommand{\gengshan}[1]{{\leavevmode\color{DarkGreen}[Gengshan: #1]}}
\newcommand{\abnote}[1]{{\leavevmode\color{DarkRed}[Andrea: #1]}}

\else
\newcommand{\gengshan}[1]{}
\newcommand{\abnote}[1]{}

\fi
\newcommand{\ourmethod}[1]{ATS}

\RequirePackage{xspace}
\makeatletter
\DeclareRobustCommand\onedot{\futurelet\@let@token\@onedot}
\def\@onedot{\ifx\@let@token.\else.\null\fi\xspace}

\def\eg{\emph{e.g}\onedot} 

\def\ie{\emph{i.e}\onedot}

\makeatother

\title{Agent-to-Sim: Learning Interactive Behavior Models from Casual Longitudinal Videos}

\author{%
  Gengshan Yang\textsuperscript{1} \quad
  Andrea Bajcsy\textsuperscript{2} \quad
  Shunsuke Saito\textsuperscript{1}$^{*}$ \quad
  Angjoo Kanazawa \textsuperscript{3}\thanks{The last two authors equally mentored this project by both having babies.} \\
    \begin{tabular}{c}
  \textsuperscript{1}Codec Avatar Labs, Meta \quad
  \textsuperscript{2}Carnegie Mellon University \quad
  \textsuperscript{3}UC Berkeley
  \end{tabular}
}

\iclrfinalcopy %
\begin{document}

\maketitle

\begin{abstract}
We present Agent-to-Sim (\ourmethod{}), a framework for learning interactive behavior models of 3D agents from casual longitudinal video collections. Different from prior works that rely on marker-based tracking and multiview cameras, \ourmethod{} learns natural behaviors of animal and human agents non-invasively through video observations recorded over a long time-span (\eg a month) in a single environment.
Modeling 3D behavior of an agent requires persistent 3D tracking (\eg, knowing which point corresponds to which) over a long time period. To obtain such data, we develop a coarse-to-fine registration method that tracks the agent and the camera over time through a canonical 3D space, resulting in a complete and persistent spacetime 4D representation. We then train a generative model of agent behaviors using paired data of perception and motion of an agent queried from the 4D reconstruction. \ourmethod{} enables real-to-sim transfer from video recordings of an agent to an interactive behavior simulator. We demonstrate results on pets (\eg, cat, dog, bunny) and human given monocular RGBD videos captured by a smartphone. Project page: \href{https://gengshan-y.github.io/agent2sim-www/}{gengshan-y.github.io/agent2sim-www/}.
\end{abstract}

\section{Introduction}
\label{sec:intro}

\begin{wrapfigure}{R}{0.27\textwidth}
\vspace{-12pt}
\centering
\includegraphics[width=0.26\textwidth]{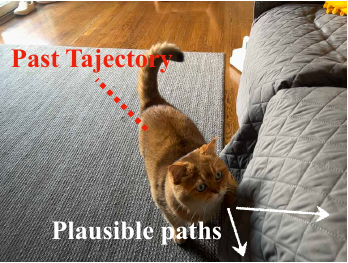}
\label{fig:front}
\vspace{-15pt}
\end{wrapfigure}

Consider an image on the right: where will the cat go and how will it move? Having seen cats interacting with the environment and people many times, we know that cats often go to the couch and follow humans around, but run away if people come too close. Our goal is to learn such a behavior model of physical agents from visual observations, just like humans can. 
This is a fundamental problem with practical application in content generation for VR/AR, robot planning in safety-critical scenarios, and behavior imitation from the real world~\citep{park2023generative, ettinger2021large, puig2023habitat, srivastava2022behavior, li2024behavior}.

\begin{figure}[t!]
\centering
\includegraphics[width=\linewidth,trim={1.5cm 5.5cm 1.5cm 5.5cm},clip]{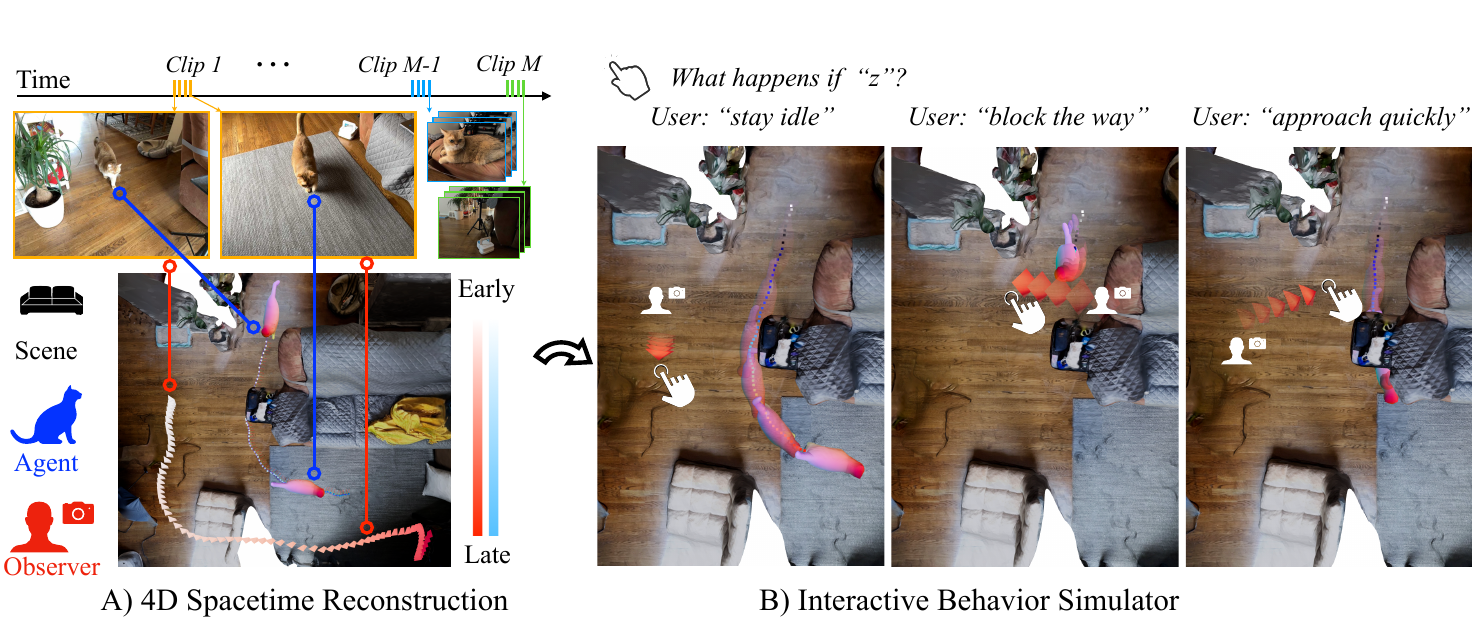}
\caption{{\bf Learning agent behavior from longitudinal casual video recordings.} We answer the following question: can we simulate the behavior of an agent, by learning from casually-captured videos of the \emph{same} agent recorded across a long period of time (\eg, a month)? A) We first reconstruct videos in 4D (3D \& time), which includes the scene, the trajectory of the agent, and the trajectory of the observer (\ie, camera held by the observer). Such individual 4D reconstructions are registered across time, resulting in a \emph{complete} and \emph{persistent} 4D representation. B) Then we learn a model of the agent for interactive behavior generation. The behavior model explicitly reasons about goals, paths, and full body movements conditioned on the agent's ego-perception and past trajectory. Such an agent representation allows generation of novel scenarios through conditioning. For example, conditioned on different observer trajectories, the cat agent chooses to walk to the carpet, stays still while quivering his tail, or hide under the tray stand. \emph{Please see videos results in the supplement}.
}
\label{fig:teaser}
\end{figure}

In a step towards building faithful models of agent behaviors, we present \ourmethod{} (Agent-to-Sim), a framework for learning interactive behavior models of 3D agents observed over a long span of time in a single environment, as shown in Fig.~\ref{fig:teaser}. %
The benefits of such a setup is multitude: 1) It is accessible, unlike approaches that capture motion data in a controlled studio with multiple cameras~\citep{mahmood2019amass, joo2017panoptic, hassan2021stochastic, kim2024parahome}, our approach only requires a single smartphone; 2) It is natural -- since the capture happens in the agent's everyday environment, it enables observing the full spectrum of natural behavior non-invasively; 3) Furthermore, it allows for longitudinal behavior capture, \eg, one that happens over a span of a month, which helps capturing a wider variety of behaviors; 4) In addition, this setup enables modeling the interactions between the agents and the observer, \ie the person taking the video.

While learning from casual longitudinal video observations has benefits, it also brings new challenges. Videos captured over time needs to be registered and reconstructed in a consistent manner. Earlier methods that reconstruct each video independently ~\citep{song2023totalrecon, gao2022monocular, park2021nerfies} is not enough, as they do not solve correspondence across the videos. In this work, we tackle a more challenging scenario: building a \emph{complete} and \emph{persistent} 4D representation from orders of magnitude more data, \eg, 20k frames of videos, and use them to learn behavior models of an agent.
To this end, we introduce a novel coarse-to-fine registration approach that re-purposes large image models, such as DiNO-v2~\citep{oquab2023dinov2}, as neural localizers, which register the cameras with respect to canonical spaces of both the agent and the scene. 
While TotalRecon~\citep{song2023totalrecon} explored reconstructing both the agent and the scene from a single video, our approach enables reconstructing multiple videos into a complete and persistent 4D representation containing the agent, the scene, and the observer. 
Then, an interactive behavior model can be learned by querying paired ego-perception and motion data from such 4D representation.

\begin{figure}[h!]
\centering
\includegraphics[width=\linewidth,trim={0cm 0cm 0cm 0cm},clip]{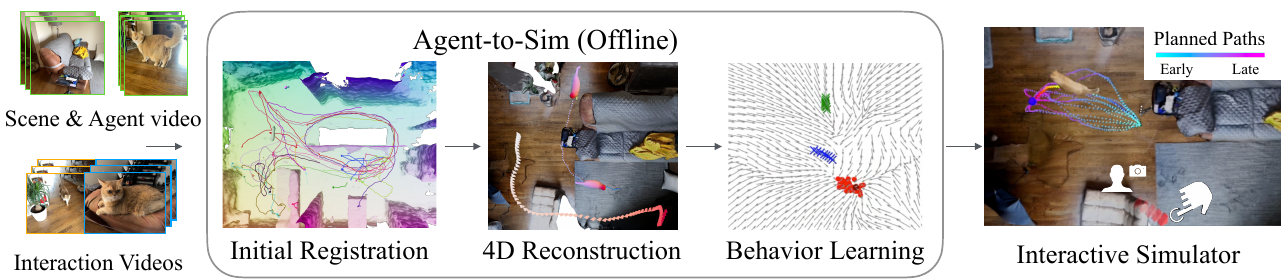}
\label{fig:framework}
\vspace{-15pt}
\end{figure}

The resulting framework, \ourmethod{}, can simulate interactive behaviors like those described at the start: agents like pets that leap onto furniture, dart quickly across the room, timidly approach nearby users, and run away if approached too quickly.  Our contributions are summarized as follows:

\begin{enumerate}
    \item \textbf{4D from Video Collections.} We build persistent and complete 4D representations from a collection of casual videos, accounting for deformations of the agent, the observer, and changes of the scene across time, enabled by a coarse-to-fine registration method.
    \item \textbf{Interactive Behavior Generation.} \ourmethod{} learns behavior that is \emph{interactive} to both the observer and 3D scene. We show results of generating plausible animal and human behaviors reactive to the observer's motion, and aware of the 3D scene.
    \item \textbf{Agent-to-Sim (ATS) Framework.} We introduce a real-to-sim framework to learn simulators of interactive agent behavior from casually-captured videos. \ourmethod{} learns natural agent behavior, and is scalable to diverse scenarios, such as animal behavior and casual events.
\end{enumerate}

\section{Related Works}

\noindent\textbf{4D Reconstruction from Monocular Videos.}
Reconstructing time-varying 3D structures from monocular videos is challenging due to its under-constrained nature. Given a monocular video, there are multiple different interpretations of the underlying 3D geometry, motion, appearance, and lighting~\citep{szeliski1997shape}. As such, previous methods often rely on category-specific 3D prior (\eg, 3D humans)~\citep{goel2023humans, SMPL:2015, kocabas2019vibe} to deal with the ambiguities. Along this line of work, there are methods to align reconstructed 3D humans to the world coordinate with the help of SLAM and visual odometry~\citep{ye2023decoupling, yuan2022glamr, kocabas2023pace}. Sitcoms3D~\citep{pavlakos2022one} reconstructs both the scene and human parameters, while relying on shot changes to determine the scale of the scene. However, the use of parametric body models limits the degrees of freedom they can capture, and makes it difficult to reconstruct agents from arbitrary categories which do not have a pre-built body model, for example, animals. Another line of work~\citep{yang2022banmo, wu2021dove} avoids using category-specific 3D priors and optimizes the shape and deformation parameters of the agent given pixel priors (\eg, optical flow and object segmentation), which works well for a broad range of categories including human, animals, and vehicles. TotalRecon~\citep{song2023totalrecon} further takes into account the background scene, such that the motion of the agent can be decoupled from the camera and aligned to the world space. However, most of the method operates on a few hundreds of frames, and none of them can reconstruct a complete 4D scene while obtaining persistent 3D tracks over orders of magnitude more data (\eg, 20k frames of videos).
We develop a coarse-to-fine registration method to register the agent and the environment into a canonical 3D space, which allows us to leverage large-scale video collection to build agent behavior models.

\noindent\textbf{Behavior Prediction and Generation.} Behavior prediction has a long history, starting from simple physics-based models such as social forces \citep{helbing1995social, alahi2016social} to more sophisticated ``planning-based'' models that cast prediction as reward optimization, where the reward is learned via inverse reinforcement learning\citep{kitani2012activity, ziebart2009planning, ma2017forecasting, ziebart2008maximum}. With the advent of large-scale motion data, generative models have been used to express behavior multi-modality~\citep{mangalam2021goals, salzmann2020trajectron++, choi2021drogon, seff2023motionlm, rhinehart2019precog}. Specifically, diffusion models are used for behavior modeling for being easily controlled via additional signals such as cost functions \citep{jiang2023motiondiffuser} or logical formulae \citep{zhong2023guided}. 
However, to capture plausible behavior of agents, they require diverse data collected in-the-wild with associated scene context, \eg, 3D map of the scene~\citep{ettinger2021large}. Such data are often manually annotated at a bounding box level~\citep{girase2021loki, ettinger2021large}, which limits the scale and the level of detail they can capture.%

\noindent\textbf{3D Agent Motion Generation.} Beyond autonomous driving setup, existing works for human and animal motion generation~\citep{tevet2022human, rempe2023trace, xie2023omnicontrol, shafir2023human, karunratanakul2023guided, pi2023hierarchical, zhang2018mode, starke2022deepphase, ling2020character, fussell2021supertrack} have been primarily using simulated data~\citep{cao2020long, van2011reciprocal} or motion capture data collected with multiple synchronized cameras~\citep{kim2024parahome, mahmood2019amass, hassan2021stochastic, luo2022artemis}. Such data provide high-quality body motion, but the interactions of the agents with the environment are either restricted to a flat ground, or a set of pre-defined furniture or objects~\citep{hassan2023synthesizing, zhao2023synthesizing, lee2023lama, zhang2023vid2player3d, menapace2024promptable}. Furthermore, the use of simulated data and motion capture data inherently limits the naturalness of the learned behavior, since agents often behave differently when being recorded in a capture studio compared to a natural environment. To bridge the gap, we develop 4D reconstruction methods to obtain high-quality trajectories of agents interacting with a natural environment, with a simple setup that can be achieved with a smartphone. %

\section{Approach}
\ourmethod{} learns behavior models of an agent in a 3D environment given RGBD videos. Sec.~\ref{sec:4d} describes our spacetime 4D representation that contains the agent, the scene, and the observer. We fit such 4D representation to a collection of videos in a coarse-to-fine manner, where the camera poses are initialized from data-driven methods and refined through differentiable rendering optimization (Sec.~\ref{sec:registration}). Given the 4D reconstruction, Sec.~\ref{sec:gdb} trains an behavior model of the agent that is \textit{interactive} to the scene and the observer. We provide a table of notations and modules in Tab.~\ref{tab:parameters}-\ref{tab:io}.

\subsection{4D Representation: Agent, Scene, and Observer}
\label{sec:4d}
Given many monocular videos, our goal is to build a complete and persistent spacetime 4D reconstruction of the underlying world, including a deformable agent, a rigid scene, and a moving observer. We factorizes the 4D reconstruction into a canonical structure and a time-varying structure. 

\noindent{\bf Canonical Structure ${\bf T=\{\sigma, c, \boldsymbol{\psi}\}}$.}
The canonical structure contains an agent neural field and a scene neural field, which are time-independent. They represent densities $\boldsymbol\sigma$, colors ${\bf c}$, and semantic features $\boldsymbol\psi$ implicitly with MLPs. To query the value at any 3D location ${\bf X}$, we have
\begin{align}
    (\sigma_{s}, {\bf c}_{s}, \boldsymbol{\psi}_{s}) &= \mathrm{MLP}_{scene}({\bf X}, \boldsymbol{\beta}_i), \label{eq:density_env}
\end{align}
\begin{align}
    (\sigma_{a}, {\bf c}_{a}, \boldsymbol{\psi}_{a}) &= \mathrm{MLP}_{agent}({\bf X}).
    \label{eq:density_agent}
\end{align}

The scene field takes in a learnable code $\boldsymbol{\beta}_i$~\citep{niemeyer2021giraffe} per-video, which can represent scenes of slightly different appearance and layout (across videos) with a shared backbone.

\noindent{\bf Time-varying Structure $\mathcal{D}=\{{\boldsymbol{\xi}}, {\bf G}, {\bf W}\}$.} The time-varying structure contains an observer and an agent. The observer is represented by the camera pose ${\boldsymbol{\xi}}_{t}\in{SE}(3)$, defined as canonical-to-camera transformations. The agent is represented by a root pose ${\bf G}^0_t\in SE(3)$, defined as canonical-to-camera transformations, and a set of 3D Gaussians, $\{{\bf G}_{t}^b\}_{\{b=1,\dots,25\}}$, referred to as ``bones''~\citep{yang2022banmo}. Bones have time-varying centers and orientations but constant scales.
Through blend-skinning~\citep{magnenat1988joint} with learned forward and backward skinning weights ${\bf W}$~\citep{saito2021scanimate}, any 3D location in the canonical space can be mapped to the time $t$ space and vice versa,
\begin{equation}\label{eq:lbs}
{\bf X}_t={\bf G}^a{\bf X} = \left(\sum_{b=1}^B {\bf W}^{b}{\bf G}^b_{t}\right){\bf X},
\end{equation}
which computes the motion of a point by blending the bone transformations (we do so in the dual quaternion space~\citep{kavan2007skinning} to ensure ${\bf G}^a$ is a valid rigid transformation). The skinning weights ${\bf W}$ are defined as the probability of a point assigned to each bone. 

\noindent{\bf Rendering.}
To render images from the 4D representation, we use differentiable volume rendering~\citep{mildenhall2020nerf} to sample rays in the camera space, map them separately to the canonical space of the scene and the agent with $\mathcal{D}$, and query values (\eg, density, color, feature) from the canonical fields of the scene and the agent. The values are then composed for ray integration~\citep{niemeyer2021giraffe}. To optimize the world representation $\{{\bf T}, \mathcal{D}\}$, we minimize the difference between the rendered pixel values and the observations, as described later in Sec.~\ref{sec:registration}. 

\subsection{Optimization: Coarse-to-fine Multi-Video Registration}
\label{sec:registration}

Given images from $M$ videos represented by color and feature descriptors~\citep{oquab2023dinov2}, $\{{\bf I}_i, {\boldsymbol\psi}_i\}_{i=\{1,\dots,M\}}$, our goal is to find a spacetime 4D representation where pixels with the same semantics can be mapped to same canonical 3D locations. Variations of appearance, lighting, and camera viewpoint across videos make it challenging to buil such persistent 4D representation.

We design a coarse-to-fine registration approach that globally aligns the agent and the observer poses to their canonical space, and then jointly optimizes the 4D representation while adjusting the poses locally. Such coarse-to-fine registration avoids bad local optima in the optimization.

\noindent{\bf Initialization: Neural Localization.} %
Due to the evolving nature of scenes across a long period of time~\citep{sun2023nothing}, there exist both global layout changes (\eg, furniture get rearranged) and appearance changes (\eg, table cloth gets replaced), making it challenging to find accurate geometric correspondences~\citep{brachmann2019ngransac, brachmann2023ace, sarlin2019coarse}. With the observation that large image models have good 3D and viewpoint awareness~\citep{el2024probing}, we adapt them for camera localization. We learn a scene-specific neural localizer that directly regresses the camera pose of an image with respect to a canonical structure,
\begin{align}\label{eq:neural_loc}
    {\boldsymbol{\xi}} = f_{\theta}(\boldsymbol{\psi}),
\end{align}
where $f_{\theta}$ is a ResNet-18~\citep{he2016deep} and $\boldsymbol{\psi}$ is the DINOv2~\citep{oquab2023dinov2} feature of the input image. We find it to be more robust than geometric correspondence, while being more computationally efficient than pairwise matches~\citep{wang2023dust3r}. To learn the neural localizer, we first capture a walk-through video and build a 3D map of the scene. Then we use it to train the neural localizer by randomly sampling camera poses ${\bf G^*}=({\bf R^*}, {\bf t^*})$ and rendering images on the fly,
\begin{align}\label{eq:fba}
    \underset{\theta}{\arg\min }\sum_j \left( \| \log({\bf R}_0^T(\theta){\bf R}^*)\| + \| {\bf t}_0(\theta) - {\bf t}^*\|_2^2  \right),
\end{align}
where we use geodesic distance~\citep{huynh2009metrics} for camera rotation and $L_2$ error for camera translation. 

Similarly, we train a camera pose estimator of the agent. First, we fit dynamic 3DGS~\citep{luiten2023dynamic, Yang_Lab4d_-_A_2023} to a long video of the agent with a complete viewpoint coverage. Then we use the dynamic 3DGS as the synthetic data generator, and train a pose regressor to predict root poses ${\bf G}^0$. During training, we randomly sample camera poses, time instances, and apply image space augmentations, including color jittering, cropping and masking.

\noindent{\bf Objective: Feature-metric Loss.}
To refine the camera registration as well as learn the deformable agent model, we fit the 4D representation $\{{\bf T}, \mathcal{D}\}$ to the data $\{{\bf I}_i, {\boldsymbol\psi}_i\}_{i=\{1,\dots,M\}}$ using differentiable rendering. Compared to fitting raw rgb values, feature descriptors from large pixel models~\citep{oquab2023dinov2} are found more robust to appearance and viewpoint changes. Therefore, we model 3D feature fields~\citep{kobayashi2022decomposing} besides colors in our canonical NeRFs (Eq.~\ref{eq:density_env}-\ref{eq:density_agent}), render them, and apply both photometric and featuremetric losses,
\begin{equation}\label{eq:loss}    
    \min_{{\bf T}, \mathcal{D}} \sum_{t}\left(\|I_t - \mathcal{R}_{I}(t; {\bf T}, \mathcal{D})\|_2^2 + \|\boldsymbol{\psi}_t - \mathcal{R}_{\boldsymbol\psi}(t; {\bf T},\mathcal{D})\|_2^2 \right) + L_{reg}({\bf T}, \mathcal{D}), 
\end{equation}
where $\mathcal{R}(\cdot)$ is the renderer described in Sec~\ref{sec:4d}. The observer (scene camera) and the agent's root pose are initialized from the coarse registration. Using featuremetric errors makes the optimization robust to change of lighting, appearance, and minor layout changes, which helps find accurate alignment across videos. We also apply a regularization term that includes eikonal loss, silhouette loss, flow loss and depth loss similar to \cite{song2023totalrecon}.

\noindent{\bf Scene Annealing.} To reconstruct a complete 3D scene when some videos are a partial capture (\eg half of the room), we encourage the reconstructed scenes across videos to be similar. To do so, we randomly {\em swap} the code $\boldsymbol\beta$ of two videos during optimization, and gradually decrease the probability of applyig swaps from  $\mathcal{P}=1.0 \rightarrow 0.05$ over the course of optimization. This regularizes the model to share structures across all videos, but keeps video-specific details (Fig.~\ref{fig:scene-registration}).

\subsection{Interactive Behavior Generation}
\label{sec:gdb}
Given the 4D representation, we extract a 3D feature volume of the scene $\boldsymbol{\Psi}$ and world-space trajectories of the observer ${\boldsymbol{\xi}}^{w}={\boldsymbol{\xi}}^{-1}$ as well as the agent ${\bf{G}}^{0,w} = \boldsymbol{\xi}^{w}{\bf G}^0, {\bf{G}}^{b,w} = {\bf G}^{0,w}\{{\bf G}^b\}_{\{b=1,\dots,25\}}$, 
as shown in Fig.~\ref{fig:ref-recon}.
Next, we learn an agent behavior model interactive with the world.

\begin{figure}[t!]
\centering
\includegraphics[width=\linewidth,trim={0.2cm 6.9cm 0.2cm 7cm},clip]{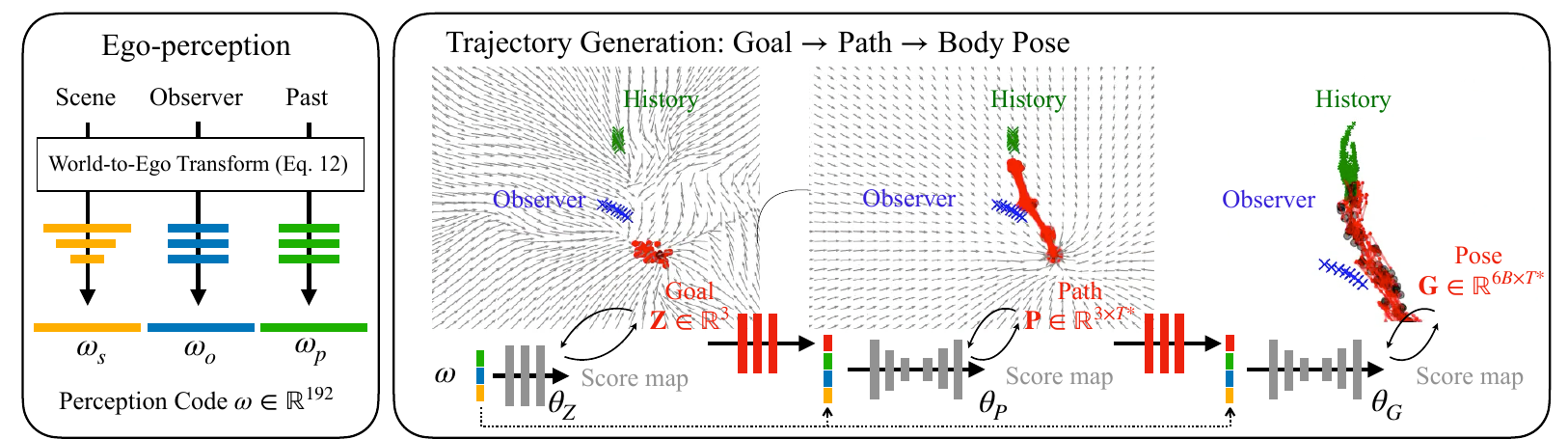}
\caption{Pipeline for behavior generation. We encode egocentric information into a perception code $\omega$, conditioned on which we generate fully body motion in a hierarchical fashion. We start by generating goals ${\bf Z}$, then paths ${\bf P}$ and finally body poses ${\bf G}$. Each node is represented by the gradient of its log distribution, trained with denoising objectives (Eq.~\ref{eq:diffusion}). Given ${\bf G}$, the full body motion of an agent can be computed via blend skinning (Eq.~\ref{eq:lbs}).}
\label{fig:motion-generation}
\end{figure}

\noindent{\bf Behavior Representation.}
We represent the behavior of an agent by its body pose in the scene space ${\bf G}\in \mathbb{R}^{6B\times T^*}$ over a time horizon $T^*=5.6$s. We design a hierarchical model as shown in Fig.~\ref{fig:motion-generation}, where the body motion ${\bf G}$ is conditioned on path ${\bf P}\in \mathbb{R}^{3 \times T^*}$, which is further conditioned on the goal ${\bf Z}\in \mathbb{R}^{3}$. Such decomposition makes it easier to learn individual components compared to learning a joint model, as shown in Tab.~\ref{tab:results} (a).

\noindent{\bf Goal Generation.} We represent a multi-modal distribution of goals ${\bf Z}\in \mathbb{R}^{3}$ by its score function $s({\bf Z,\sigma})\in\mathbb{R}^{3}$~\citep{ho2020denoising, song2020score}. The score function is implemented as an MLP,
\begin{align}
    s({\bf Z}; \sigma) = \mathrm{MLP}_{\theta_{\bf Z}}({\bf Z}, \sigma),
\end{align}
trained by predicting the amount of noise $\boldsymbol\epsilon$ added to the clean goal, given the corrupted goal ${\bf Z}+\boldsymbol{\epsilon}$:
\begin{align}
\label{eq:diffusion}
    \underset{\theta_{\bf Z}}{\arg\min } \mathbb{E}_{\boldsymbol{Z}} \mathbb{E}_{\sigma \sim q(\sigma)} \mathbb{E}_{\boldsymbol{\epsilon} \sim \mathcal{N}\left(\mathbf{0}, \sigma^2\boldsymbol{I}\right)}\left\|{\mathrm{MLP}}_{\theta_{\bf Z}}(\boldsymbol{Z}+\boldsymbol{\epsilon} ; \sigma)-\boldsymbol{\epsilon}\right\|_2^2.
\end{align}

\noindent{\bf Trajectory Generation.} 
To generate path conditioned on goals, we represent its score function as
\begin{align}
    s({\bf P}; \sigma) = \mathrm{ControlUNet}_{\theta_{\bf P}}({\bf P}, {\bf Z}, \sigma) ,
\end{align}
where the Control UNet contains two standard UNets with the same architecture~\citep{zhang2023adding, xie2023omnicontrol}, one taking $({\bf P}, \sigma)$ as input to perform unconditional generation, another taking $({\bf Z}, \sigma)$ as inputs to inject goal conditions densely into the neural network blocks of the first one. We apply the same architecture to generate body poses conditioned on paths,
\begin{align}
    s({\bf G}; \sigma) = \mathrm{ControlUNet}_{\theta_{\bf G}}({\bf G}, {\bf P}, \sigma).
\end{align}
Compared to concatenating the goal condition to the noise latent, this encourages close alignment between the input goal and the path~\citep{xie2023omnicontrol}. 

\noindent{\bf Ego-Perception of the World.} To generate plausible interactive behaviors, we encode the world \textit{egocentrically} perceived by the agent, and use it to condition the behavior generation. The ego-perception code $\omega$ contains a scene code $\omega_s$, an observer code $\omega_o$, and a past code $\omega_p$, as detailed later. The ego-perception code is concatenated to the noise value $\sigma$ and passed to the denoising networks. Transforming the world to the egocentric coordinates avoids over-fitting to specific locations of the scene (Tab.~\ref{tab:results}-(b)). We find that a specific behavior can be learned and generalized to novel situations even when seen once. Although there's only one data point where the cat jumps off the dining table, our method can generate diverse motion of cat jumping off the table while landing at different locations (to the left, middle, and right of the table). Please see Fig.~\ref{fig:generalization} for the corresponding visual. 

\noindent{\bf Scene, Observer, and Past Encoding.}
To encode the scene, we extract a latent representation from a local feature volume around the agent, where the volume is queried from the 3D feature volume by transforming the sampled ego-coordinates ${\bf X}^{a}$ using the agent-to-world transformation at time $t$,
\begin{align}
\omega_{s} = \mathrm{ResNet3D}_{\theta_\psi}(\boldsymbol\Psi_{s}({\bf X}_w)), \quad {\bf X}^{w} = ({\bf G}^{0,w}_t) {\bf X}^{a}.
\end{align}
where $\mathrm{ResNet3D}_{\theta_\phi}$ is a 3D ConvNet with residual connections, and $\omega_s\in\mathbb{R}^{64}$. 

To encode the observer's motion in the past $T'=0.8$s seconds, we transform observer's trajectories to the ego-coordinate,
\begin{align}
    \omega_{o} = \mathrm{MLP}_{\theta_{o}}({\boldsymbol{\xi}^{a}}), \quad \boldsymbol{\xi}^{a} = ({\bf G}^{0,w}_t)^{-1} \boldsymbol{\xi}^w,
\end{align}
where $\omega_{o}\in\mathbb{R}^{64}$ represents the observer perceived by the agent. Accounting for the external factors from the ``world'' enables interactive behavior generation, where the motion of an agent follows the environment constraints and is influenced by the trajectory of the observer, as shown in Fig.~\ref{fig:condition-signal}. 

We additionally encode the root and body motion of the agent in the past $T'$ seconds,
\begin{align}
    \omega_{p} = \mathrm{MLP}_{\theta_{p}}({{\bf G}^{\{0,\dots,B\},a}}), \quad {\bf G}^{\{0,\dots,B\},a} = ({\bf G}^{0,w}_t)^{-1} {\bf G}^{\{0,\dots,B\}, w}.
\end{align}
By conditioning on the past motion, we can generate long sequences by chaining individual ones.

\section{Experiments}
\noindent\textbf{Dataset.}
We collect a dataset that emphasizes interactions of an agent with the environment and the observer. As shown in Tab.~\ref{tab:dataset}, it contains RGBD iPhone video collections of 4 agents in 3 different scenes, where human and cat share the same scene. The dataset is curated to contain diverse motion of agents, including walking, lying down, eating, as well as diverse interaction patterns with the environment, including following the camera, sitting on a coach, etc.

\begin{figure}[t!]
\centering
\includegraphics[width=\linewidth,trim={0cm 7cm 0cm 8cm},clip]{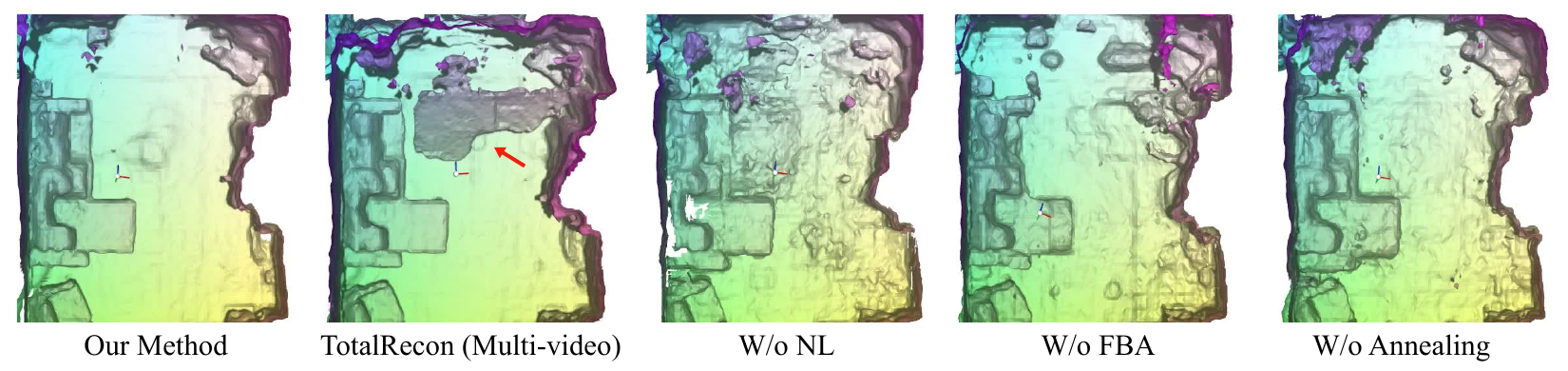}
\caption{{\bf Comparison on multi-video scene reconstruction}. We show birds-eye-view rendering of the reconstructed scene using the bunny dataset. 
Compared to TotalRecon that does not register multiple videos, \ourmethod{} produces higher-quality scene reconstruction. Neural localizer (NL) and featuremetric losses (FBA) are shown important for camera registration. Scene annealing is important for reconstructing a complete scene from partial video captures.}
\label{fig:scene-registration}
\end{figure}

\subsection{4D Reconstruction of Agent \& Environment }
\noindent\textbf{Implementation Details.} We take a video collection of the same agent as input, and build a 4D reconstruction of the agent, the scene, and the observer. We extract frames from the videos at 10 FPS, and use off-the-shelf models to produce augmented image measurements, including object segmentation~\citep{yang2023track}, optical flow~\citep{yang2019volumetric}, DINOv2 features~\citep{oquab2023dinov2}. We use AdamW to first optimize the environment with feature-metric loss for 30k iterations, and then jointly optimize the environment and agent for another 30k iterations with all losses in Eq.~\ref{eq:loss}. Optimization takes roughly 24 hours. 8 A100 GPUs are used to optimize 23 videos of the cat data, and 1 A100 GPU is used in a 2-3 video setup (for dog, bunny, and human). 

\begin{table*}[!t]
    \centering
    \begin{minipage}[t]{0.56\textwidth}
        \caption{\textbf{Evaluation of Camera Registration}.}
        \small
        \centering
        \resizebox{\textwidth}{!}{%
        \begin{tabular}{lccc}
        \toprule
        Method & Rotation Error (\textdegree) & Translation Error (m) \\
        \midrule
        Ours & {\bf 6.35} & {\bf 0.41} \\
        w/o Neural Localizer & 37.59 & 0.83 \\
        w/o Featuremetric BA & 22.47 & 1.30 \\
        Multi-video TotalRecon & 59.19 & 0.68 \\
        \bottomrule
        \end{tabular}}
        \label{tab:camera_registration}
    \end{minipage}%
    \hfill
    \begin{minipage}[t]{0.43\textwidth}
        \caption{\textbf{Dataset used in \ourmethod{}}.}
        \small
        \centering
        \resizebox{\textwidth}{!}{%
        \begin{tabular}{lrrrr}
        \toprule
         & Videos & \ Length & Unique Days / Span \\
        \midrule
        Cat & 23 & 25m 39s & 9 / 37 days\\
        Human & 5 & 9m 27s & 2 / 4 days\\
        Dog & 3 & 7m 13s & 1 / 1 day\\
        Bunny & 2 & 1m 48s & 1 / 1 day\\
        \bottomrule
        \end{tabular}}
        \label{tab:dataset}
    \end{minipage}
\end{table*}

\noindent\textbf{Results of Camera Registration.} 
We evaluate camera registration using GT cameras estimated from annotated 2D correspondences. A visual of the annotated correspondence and 3D alignment can be found in Fig.~\ref{fig:corresp-anno}. We report camera translation and rotation errors in Tab.~\ref{tab:camera_registration}. We observe that removing neural localization (Eq.~\ref{eq:neural_loc}) produces significantly larger localization error (\eg, Rotation error: 6.35 vs 37.56). Removing feature-metric bundle adjustment (Eq.~\ref{eq:fba}) also increases the error (\eg, Rotation error: 6.35 vs 22.47). Our method outperforms multi-video TotalRecon by a large margin due to the above innovations.

A visual comparison on scene registration is shown in Fig.~\ref{fig:scene-registration}. Without the ability to register multiple videos, TotalRecon produces protruded and misaligned structures (as pointed by the red arrow). In contrast, our method reconstructs a single coherent scene. With featuremetric alignment (FBA) alone but without a good camera initialization from neural localization (NL), our method produces inaccurate reconstruction due to inaccurate global alignment in cameras poses. Removing FBA while keeping NL, the method fails to accurately localize the cameras and produces noisy scene structures. Finally, removing scene annealing procures lower quality reconstruction due to the partial capture.

\noindent\textbf{Results of 4D Reconstruction.} 
We evaluate the accuracy of 4D reconstruction using synchronized videos captured with two moving iPhone cameras looking from opposite views. The results can be found in Tab.~\ref{tab:4dreconresults}. We compute the GT relative camera pose between the two cameras from 2D correspondence annotations. One of the synchronized videos is used for 4D reconstruction, and the other one is used as held-out test data. For evaluation, we render novel views from the held-out cameras and compute novel view depth accuracy DepthAcc (depth accuracy thresholded at 0.1m) for all pixels, agent, and scene, following TotalRecon. Our method outperforms both the multi-video and single-video versions of TotalRecon by a large margin in terms of depth accuracy and LPIPS, due to the ability of leveraging multiple videos. Please see Fig.~\ref{fig:4d-eval} for the corresponding visual.

\begin{table}[!t]
    \caption{\textbf{Evaluation of 4D Reconstruction}. SV: Single-video. MV: Multi-video.}
    \small
    \centering
    \resizebox{\columnwidth}{!}{%
    \begin{tabular}{lcccccc}
    \toprule
    Method & DepthAcc (all) & DepthAcc (fg) & DepthAcc (bg) & LPIPS (all) & LPIPS (fg) & LPIPS (bg) \\
    \midrule
    Ours & \textbf{0.708} & \textbf{0.695} & \textbf{0.703} & \textbf{0.613} & \textbf{0.609} & \textbf{0.613} \\
    SV TotalRecon & 0.533 & 0.685 & 0.518 & 0.641 & 0.619 & 0.641 \\ 
    MV TotalRecon & 0.099 & 0.647 & 0.053 & 0.634 & 0.666 & 0.633 \\ 
    \bottomrule
    \end{tabular}}
    \label{tab:4dreconresults}
\end{table}

\begin{figure}[h]
\centering
\includegraphics[width=1\linewidth,trim={6.5cm 2cm 6.5cm 2cm},clip]{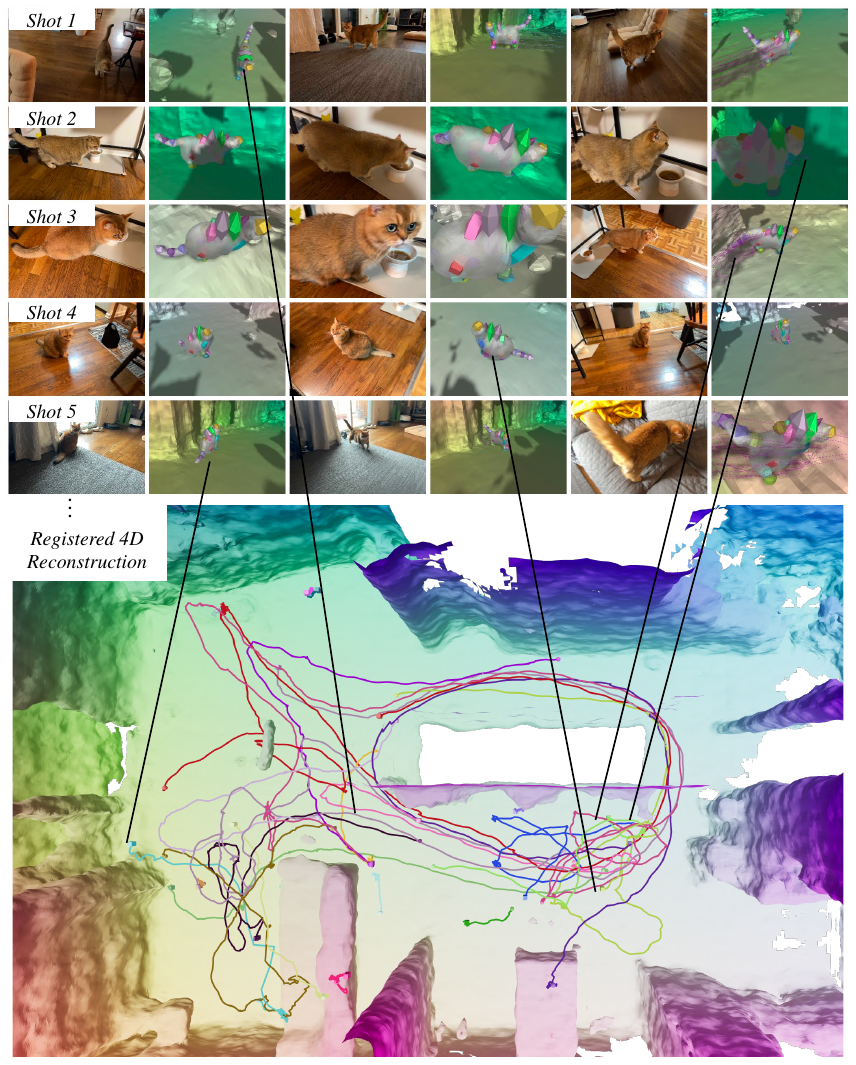}
\caption{{\bf Results of 4D reconstruction}. Top: reference images and renderings. Background color represents correspondence. Colored blobs on the cat represent $B=25$ bones (\eg, head is represented by the yellow blob).
The magenta colored lines represents reconstructed trajectories of each blob in the world space. 
Bottom: Bird's eye view of the reconstructed scene and agent trajectories, registered to the same scene coordinate. Each colored line represents a unique video sequence where boxes and spheres indicate the starting and the end location.
}
\label{fig:ref-recon}
\end{figure}

Qualitative results of 4D reconstruction can be found in Fig.~\ref{fig:ref-recon} and the supplementary webpage. A visual comparison with TotalRecon (Single Video) is shown in Fig.~\ref{fig:vstotalrecon}, where we show that multiple videos helps improving the reconstruction quality on both the agent and the scene.

\subsection{Interactive Agent Behavior Prediction}

\noindent\textbf{Dataset.}
We train agent-specific behavior models for cat, dog, bunny, and human using 4D reconstruction from their corresponding video collections. We use the cat dataset for quantitative evaluation, where the data are split into a training set of 22 videos and a test set of 1 video.

\noindent\textbf{Implementation Details.} Our model consists of three diffusion models, for goal, path, and full body motion respectively. To train the behavior model, we slice the reconstructed trajectory in the training set into overlapping window of $6.4$s, resulting in 12k data samples. We use AdamW to optimize the parameters of the scores functions $\{{\theta_{\bf Z}}, {\theta_{\bf P}}, {\theta_{\bf G}}\}$ and the ego-perception encoders $\{{\theta_{{\psi}}}, {\theta_{o}}, {\theta_{p}}\}$ for 120k steps with batch size 1024. Training takes 10 hours on a single A100 GPU. Each diffusion model is trained with random dropout of the conditioning~\citep{ho2022classifier}. 

\noindent\textbf{Metrics.}
The behavior of an agent can be evaluated along multiple axes, and we focus on goal, path, and body motion prediction. For goal prediction, we use minimum displacement error (minDE)~\citep{chai2019multipath}. The evaluation asks the model to produce $K=16$ hypotheses, and minDE finds the one closest to the ground-truth to compute the distance. For path and body motion prediction, we use minimum average displacement error (minADE), which are similar to goal prediction, but additionally averages the distance over path and joint angles before taking the min. When evaluating path prediction and body motion prediction, the output is conditioned on the ground-truth goal and path respectively.

\begin{table}[!t]
    \caption{\textbf{End-to-end Evaluation of Interactive Behavior Prediction.} We report results of predicting goal, path, orientation, and joint angles, using $K=16$ samples across $L=12$ trials. The metrics are minimum average displacement error (minADE) with standard deviations ($\pm\sigma$). The lower the better and the best results are in bold.}
    \small
    \centering
    \resizebox{\columnwidth}{!}{%
    \begin{tabular}{lcccc}
    \toprule
    Method & Goal (m) $\downarrow$ & Path (m) $\downarrow$& Orientation ($\mathrm{rad}$) $\downarrow$& Joint Angles ($\mathrm{rad}$)$\downarrow$\\
    \midrule
    Location prior~\citep{ziebart2009planning} 
    & 0.663$^{\pm0.307}$ & N.A. & N.A. & N.A. \\
    Gaussian~\citep{kendall2017uncertainties} 
    &   0.942$^{\pm  0.081}$ &   0.440 $^{\pm  0.002}$ &   1.099 $^{\pm  0.003}$ &   0.295 $^{\pm  0.001}$ \\
    \ourmethod{} (Ours) 
    &   {\bf 0.448$^{\pm  0.146}$} &   {\bf 0.234} $^{\pm  0.054}$ &   {\bf 0.550} $^{\pm  0.112}$ &   {\bf 0.237} $^{\pm  0.006}$ \\
    \midrule
    (a) hier$\rightarrow$1-stage~\citep{tevet2022human}
    &   1.322$^{\pm  0.071}$ &   0.575 $^{\pm  0.026}$ &   0.879 $^{\pm  0.041}$ &   0.263 $^{\pm  0.007}$ \\
    (b) ego$\rightarrow$world~\citep{rhinehart2016learning} 
    &   1.164$^{\pm  0.043}$ &   0.577 $^{\pm  0.022}$ &   0.873 $^{\pm  0.027}$ &   0.295 $^{\pm  0.006}$ \\
    (c) w/o observer $\omega_o$ 
    &   0.647$^{\pm  0.148}$ &   0.327 $^{\pm  0.076}$ &   0.620 $^{\pm  0.092}$ &   0.240 $^{\pm  0.006}$ \\
    (d) w/o scene $\omega_{s}$ 
    &   0.784$^{\pm  0.126}$ &   0.340 $^{\pm  0.051}$ &   0.678 $^{\pm  0.081}$ &   0.243 $^{\pm  0.007}$ \\
    \bottomrule
    \end{tabular}}
    \label{tab:results}
\end{table}

\begin{table}[!t]
    \caption{\textbf{Evaluation of Spatial Control.} We evaluate goal-conditioned path generation and path-conditoned full body motion generation respectively. %
    }
    \small
    \centering
    \resizebox{0.9\columnwidth}{!}{%
    \begin{tabular}{lcccc}
    \toprule
    Method &  Path (m) $\downarrow$& Orientation ($\mathrm{rad}$) $\downarrow$& Joint Angles ($\mathrm{rad}$)$\downarrow$\\
    \midrule
    Gaussian~\citep{kendall2017uncertainties} 
    & 0.206$^{\pm0.002}$ & 0.370$^{\pm0.003}$ & 0.232$^{\pm0.001}$ \\
    \ourmethod{} (Ours) 
    &\bf{0.115}$^{\pm0.006}$ & \bf{0.331}$^{\pm0.004}$ & \bf{0.213}$^{\pm0.001}$ \\
    \midrule
    (a) ego$\rightarrow$world~\citep{rhinehart2016learning} 
    & 0.209$^{\pm0.002}$ & 0.429$^{\pm0.006}$ & 0.250$^{\pm0.002}$ \\
    (b) control-unet$\rightarrow$code
    &   0.146 $^{\pm  0.005}$ &   0.351 $^{\pm  0.004}$ &   0.220 $^{\pm  0.001}$ \\
    \bottomrule
    \end{tabular}}
    \label{tab:results-b}
\end{table}

\begin{figure}[t!]
\centering
\includegraphics[width=\linewidth,trim={0cm 5cm 0cm 5cm},clip]{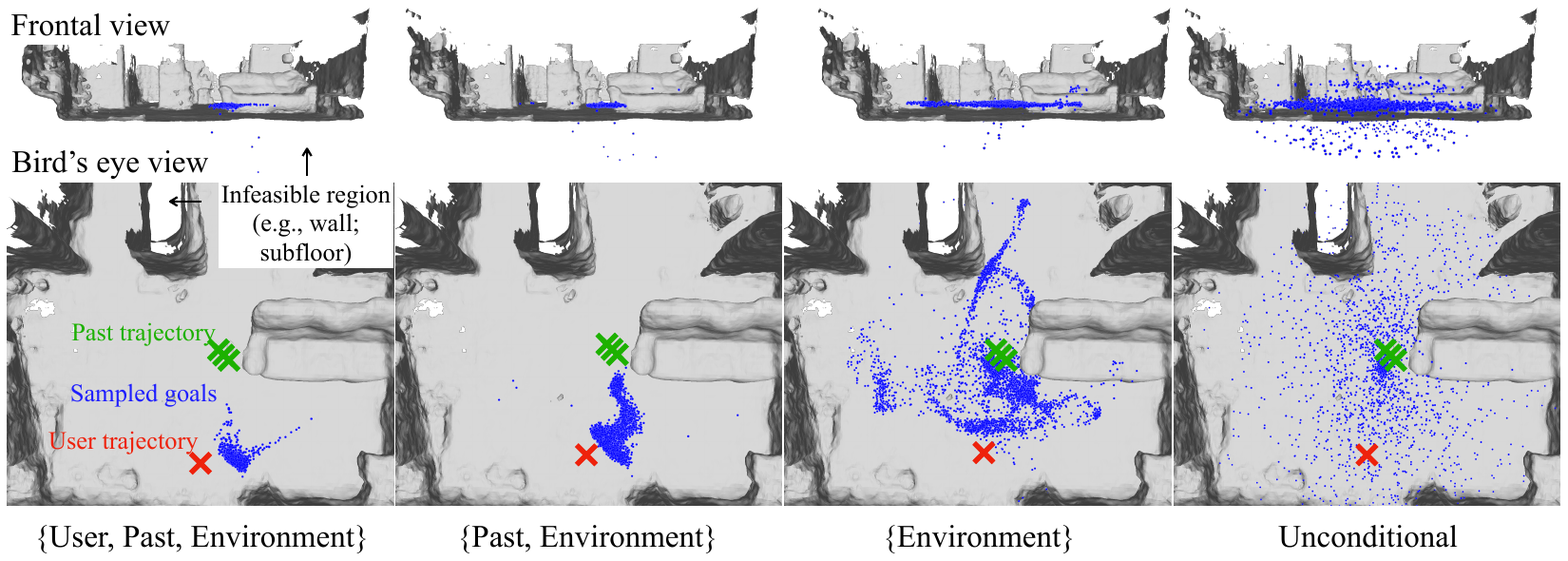}
\caption{Analysis of conditioning signals. We show results of removing one conditioning signal at a time. Removing observer conditioning and past trajectory conditioning makes the sampled goals more spread out (\eg, regions both in front of the agent and behind the agent); removing the environment conditioning introduces infeasible goals that penetrate the ground and the walls.}
\label{fig:condition-signal}
\end{figure}

\noindent{\bf Comparisons and Ablations.} We compare to related methods in our setup and the quantitative results are shown in Tab.~\ref{tab:results}. To predict the goal of an agent, classic methods build statistical models of how likely an agent visits a spatial location of the scene, referred to as location prior~\citep{ziebart2009planning, kitani2012activity}. Given the extracted 3D trajectories of an agent in the egocentric coordinate, we build a 3D preference map over 3D locations as a histogram, which can be turned into probabilities and used to sample goals. Since it does not take into account of the scene and the observer, it fails to accurately predict the goal. We implement a ``Gaussian'' baseline that represents the goal, path, and full body motion as Gaussians, by predicting both the mean and variance of Gaussian distributions~\citep{kendall2017uncertainties}. It is trained on the same data and takes the same input as \ourmethod{}. As a result, the ``Gaussian'' baseline performs worse than \ourmethod{} since Gaussian cannot represent multi-modal distributions of agent behaviors, resulting in mode averaging. We implement a 1-stage model similar to MDM~\citep{tevet2022human} that directly denoises body motion without predicting goals and paths (Tab.~\ref{tab:results} (a)). Our hierarchical model out-performs 1-stage by a large margin. We posit hierarchical model makes it easier to learn individual modules. Finally, learning behavior in the world coordinates (Tab.~\ref{tab:results} (b)), akin to ActionMap~\citep{rhinehart2016learning}, performs worse for all metrics due to the over-fits to specific locations of the scene.

\noindent\textbf{Analysing Interactions.}
We analyse the agent's interactions with the environment and the observer by removing the conditioning signals and study their influence on behavior prediction. In Fig.~\ref{fig:condition-signal}, we show that by gradually removing conditional signals, the generated goal samples become more spread out. In Tab.~\ref{tab:results}, we drop one of the conditioning signals at a time, and find that dropping either the observer conditioning or the environment conditioning increases behavior prediction errors. 

\noindent\textbf{Spatial Control.}
Besides generating behaviors conditioned on agent's perception, we could also condition on user-provided spatial signals (\eg, goal and path) to steer the generated behavior. The results are reported in Tab.~\ref{tab:results-b}. We found that \ourmethod{} performs better than ``Gaussians'' for behavior control due to its ability to represent complex distributions. Furthermore, egocentric representation produces better behavior generation results. Finally, replacing control-unet architecture by concatenating spatial control with perception codes produces worse alignment (\eg, Path error: 0.115 vs 0.146).

\section{Conclusion}
We have presented a method for learning interactive behavior of agents grounded in 3D environments. Given multiple casually-captured video recordings, we build persistent 4D reconstructions including the agent, the environment, and the observer. Such data collected over a long time period allows us to learn a behavior model of the agent that is reactive to the observer and respects the environment constraints. We validate our design choices on casual video collections, and show better results than prior work for 4D reconstruction and interactive behavior prediction.

\subsubsection*{Acknowledgments}
This project was funded in part by NSF:CNS-2235013 and IARPA DOI/IBC No. 140D0423C0035

\clearpage
\newpage
\bibliography{iclr2025_conference}

\begin{thebibliography}{83}
\providecommand{\natexlab}[1]{#1}
\providecommand{\url}[1]{\texttt{#1}}
\expandafter\ifx\csname urlstyle\endcsname\relax
  \providecommand{\doi}[1]{doi: #1}\else
  \providecommand{\doi}{doi: \begingroup \urlstyle{rm}\Url}\fi

\bibitem[Alahi et~al.(2016)Alahi, Goel, Ramanathan, Robicquet, Fei-Fei, and Savarese]{alahi2016social}
Alexandre Alahi, Kratarth Goel, Vignesh Ramanathan, Alexandre Robicquet, Li~Fei-Fei, and Silvio Savarese.
\newblock Social lstm: Human trajectory prediction in crowded spaces.
\newblock In \emph{CVPR}, pp.\  961--971, 2016.

\bibitem[Brachmann \& Rother(2019)Brachmann and Rother]{brachmann2019ngransac}
Eric Brachmann and Carsten Rother.
\newblock {N}eural- {G}uided {RANSAC}: {L}earning where to sample model hypotheses.
\newblock In \emph{ICCV}, 2019.

\bibitem[Brachmann et~al.(2023)Brachmann, Cavallari, and Prisacariu]{brachmann2023ace}
Eric Brachmann, Tommaso Cavallari, and Victor~Adrian Prisacariu.
\newblock Accelerated coordinate encoding: Learning to relocalize in minutes using rgb and poses.
\newblock In \emph{CVPR}, 2023.

\bibitem[Cao et~al.(2020)Cao, Gao, Mangalam, Cai, Vo, and Malik]{cao2020long}
Zhe Cao, Hang Gao, Karttikeya Mangalam, Qi-Zhi Cai, Minh Vo, and Jitendra Malik.
\newblock Long-term human motion prediction with scene context.
\newblock In \emph{ECCV}, pp.\  387--404. Springer, 2020.

\bibitem[Chai et~al.(2019)Chai, Sapp, Bansal, and Anguelov]{chai2019multipath}
Yuning Chai, Benjamin Sapp, Mayank Bansal, and Dragomir Anguelov.
\newblock Multipath: Multiple probabilistic anchor trajectory hypotheses for behavior prediction.
\newblock \emph{arXiv preprint arXiv:1910.05449}, 2019.

\bibitem[Choi et~al.(2021)Choi, Malla, Patil, and Choi]{choi2021drogon}
Chiho Choi, Srikanth Malla, Abhishek Patil, and Joon~Hee Choi.
\newblock Drogon: A trajectory prediction model based on intention-conditioned behavior reasoning.
\newblock In \emph{CoRL}, pp.\  49--63. PMLR, 2021.

\bibitem[El~Banani et~al.(2024)El~Banani, Raj, Maninis, Kar, Li, Rubinstein, Sun, Guibas, Johnson, and Jampani]{el2024probing}
Mohamed El~Banani, Amit Raj, Kevis-Kokitsi Maninis, Abhishek Kar, Yuanzhen Li, Michael Rubinstein, Deqing Sun, Leonidas Guibas, Justin Johnson, and Varun Jampani.
\newblock Probing the 3d awareness of visual foundation models.
\newblock In \emph{CVPR}, pp.\  21795--21806, 2024.

\bibitem[Ettinger et~al.(2021)Ettinger, Cheng, Caine, Liu, Zhao, Pradhan, Chai, Sapp, Qi, Zhou, et~al.]{ettinger2021large}
Scott Ettinger, Shuyang Cheng, Benjamin Caine, Chenxi Liu, Hang Zhao, Sabeek Pradhan, Yuning Chai, Ben Sapp, Charles~R Qi, Yin Zhou, et~al.
\newblock Large scale interactive motion forecasting for autonomous driving: The waymo open motion dataset.
\newblock In \emph{ICCV}, pp.\  9710--9719, 2021.

\bibitem[Fussell et~al.(2021)Fussell, Bergamin, and Holden]{fussell2021supertrack}
Levi Fussell, Kevin Bergamin, and Daniel Holden.
\newblock Supertrack: Motion tracking for physically simulated characters using supervised learning.
\newblock \emph{ACM Transactions on Graphics (TOG)}, 40\penalty0 (6):\penalty0 1--13, 2021.

\bibitem[Gao et~al.(2022)Gao, Li, Tulsiani, Russell, and Kanazawa]{gao2022monocular}
Hang Gao, Ruilong Li, Shubham Tulsiani, Bryan Russell, and Angjoo Kanazawa.
\newblock Monocular dynamic view synthesis: A reality check.
\newblock \emph{NeurIPS}, 35:\penalty0 33768--33780, 2022.

\bibitem[Girase et~al.(2021)Girase, Gang, Malla, Li, Kanehara, Mangalam, and Choi]{girase2021loki}
Harshayu Girase, Haiming Gang, Srikanth Malla, Jiachen Li, Akira Kanehara, Karttikeya Mangalam, and Chiho Choi.
\newblock Loki: Long term and key intentions for trajectory prediction.
\newblock In \emph{ICCV}, pp.\  9803--9812, 2021.

\bibitem[Goel et~al.(2023)Goel, Pavlakos, Rajasegaran, Kanazawa*, and Malik*]{goel2023humans}
Shubham Goel, Georgios Pavlakos, Jathushan Rajasegaran, Angjoo Kanazawa*, and Jitendra Malik*.
\newblock Humans in 4{D}: Reconstructing and tracking humans with transformers.
\newblock In \emph{ICCV}, 2023.

\bibitem[Hassan et~al.(2021)Hassan, Ceylan, Villegas, Saito, Yang, Zhou, and Black]{hassan2021stochastic}
Mohamed Hassan, Duygu Ceylan, Ruben Villegas, Jun Saito, Jimei Yang, Yi~Zhou, and Michael~J Black.
\newblock Stochastic scene-aware motion prediction.
\newblock In \emph{ICCV}, pp.\  11374--11384, 2021.

\bibitem[Hassan et~al.(2023)Hassan, Guo, Wang, Black, Fidler, and Peng]{hassan2023synthesizing}
Mohamed Hassan, Yunrong Guo, Tingwu Wang, Michael Black, Sanja Fidler, and Xue~Bin Peng.
\newblock Synthesizing physical character-scene interactions.
\newblock In \emph{SIGGRAPH 2023 Conference Proceedings}, pp.\  1--9, 2023.

\bibitem[He et~al.(2016)He, Zhang, Ren, and Sun]{he2016deep}
Kaiming He, Xiangyu Zhang, Shaoqing Ren, and Jian Sun.
\newblock Deep residual learning for image recognition.
\newblock In \emph{CVPR}, pp.\  770--778, 2016.

\bibitem[Helbing \& Molnar(1995)Helbing and Molnar]{helbing1995social}
Dirk Helbing and Peter Molnar.
\newblock Social force model for pedestrian dynamics.
\newblock \emph{Physical review E}, 51\penalty0 (5):\penalty0 4282, 1995.

\bibitem[Ho \& Salimans(2022)Ho and Salimans]{ho2022classifier}
Jonathan Ho and Tim Salimans.
\newblock Classifier-free diffusion guidance.
\newblock \emph{arXiv preprint arXiv:2207.12598}, 2022.

\bibitem[Ho et~al.(2020)Ho, Jain, and Abbeel]{ho2020denoising}
Jonathan Ho, Ajay Jain, and Pieter Abbeel.
\newblock Denoising diffusion probabilistic models.
\newblock \emph{NeurIPS}, 33:\penalty0 6840--6851, 2020.

\bibitem[Huynh(2009)]{huynh2009metrics}
Du~Q Huynh.
\newblock Metrics for 3d rotations: Comparison and analysis.
\newblock \emph{Journal of Mathematical Imaging and Vision}, 35:\penalty0 155--164, 2009.

\bibitem[Jiang et~al.(2023)Jiang, Cornman, Park, Sapp, Zhou, Anguelov, et~al.]{jiang2023motiondiffuser}
Chiyu Jiang, Andre Cornman, Cheolho Park, Benjamin Sapp, Yin Zhou, Dragomir Anguelov, et~al.
\newblock Motiondiffuser: Controllable multi-agent motion prediction using diffusion.
\newblock In \emph{CVPR}, pp.\  9644--9653, 2023.

\bibitem[Joo et~al.(2017)Joo, Simon, Li, Liu, Tan, Gui, Banerjee, Godisart, Nabbe, Matthews, et~al.]{joo2017panoptic}
Hanbyul Joo, Tomas Simon, Xulong Li, Hao Liu, Lei Tan, Lin Gui, Sean Banerjee, Timothy Godisart, Bart Nabbe, Iain Matthews, et~al.
\newblock Panoptic studio: A massively multiview system for social interaction capture.
\newblock \emph{TPAMI}, 41\penalty0 (1):\penalty0 190--204, 2017.

\bibitem[Karunratanakul et~al.(2023)Karunratanakul, Preechakul, Suwajanakorn, and Tang]{karunratanakul2023guided}
Korrawe Karunratanakul, Konpat Preechakul, Supasorn Suwajanakorn, and Siyu Tang.
\newblock Guided motion diffusion for controllable human motion synthesis.
\newblock In \emph{ICCV}, pp.\  2151--2162, 2023.

\bibitem[Kavan et~al.(2007)Kavan, Collins, {\v{Z}}{\'a}ra, and O'Sullivan]{kavan2007skinning}
Ladislav Kavan, Steven Collins, Ji{\v{r}}{\'\i} {\v{Z}}{\'a}ra, and Carol O'Sullivan.
\newblock Skinning with dual quaternions.
\newblock In \emph{Proceedings of the 2007 symposium on Interactive 3D graphics and games}, pp.\  39--46, 2007.

\bibitem[Kendall \& Gal(2017)Kendall and Gal]{kendall2017uncertainties}
Alex Kendall and Yarin Gal.
\newblock What uncertainties do we need in bayesian deep learning for computer vision?
\newblock In \emph{NIPS}, 2017.

\bibitem[Kerbl et~al.(2023)Kerbl, Kopanas, Leimk{\"u}hler, and Drettakis]{kerbl20233d}
Bernhard Kerbl, Georgios Kopanas, Thomas Leimk{\"u}hler, and George Drettakis.
\newblock 3d gaussian splatting for real-time radiance field rendering.
\newblock \emph{ACM Transactions on Graphics}, 42\penalty0 (4):\penalty0 1--14, 2023.

\bibitem[Kim et~al.(2024)Kim, Kim, Na, and Joo]{kim2024parahome}
Jeonghwan Kim, Jisoo Kim, Jeonghyeon Na, and Hanbyul Joo.
\newblock Parahome: Parameterizing everyday home activities towards 3d generative modeling of human-object interactions.
\newblock \emph{arXiv preprint arXiv:2401.10232}, 2024.

\bibitem[Kitani et~al.(2012)Kitani, Ziebart, Bagnell, and Hebert]{kitani2012activity}
Kris~M Kitani, Brian~D Ziebart, James~Andrew Bagnell, and Martial Hebert.
\newblock Activity forecasting.
\newblock In \emph{ECCV}, pp.\  201--214. Springer, 2012.

\bibitem[Kobayashi et~al.(2022)Kobayashi, Matsumoto, and Sitzmann]{kobayashi2022decomposing}
Sosuke Kobayashi, Eiichi Matsumoto, and Vincent Sitzmann.
\newblock Decomposing nerf for editing via feature field distillation.
\newblock \emph{Advances in Neural Information Processing Systems}, 35:\penalty0 23311--23330, 2022.

\bibitem[Kocabas et~al.(2020)Kocabas, Athanasiou, and Black]{kocabas2019vibe}
Muhammed Kocabas, Nikos Athanasiou, and Michael~J. Black.
\newblock Vibe: Video inference for human body pose and shape estimation.
\newblock In \emph{CVPR}, June 2020.

\bibitem[Kocabas et~al.(2023)Kocabas, Yuan, Molchanov, Guo, Black, Hilliges, Kautz, and Iqbal]{kocabas2023pace}
Muhammed Kocabas, Ye~Yuan, Pavlo Molchanov, Yunrong Guo, Michael~J Black, Otmar Hilliges, Jan Kautz, and Umar Iqbal.
\newblock Pace: Human and camera motion estimation from in-the-wild videos.
\newblock \emph{arXiv preprint arXiv:2310.13768}, 2023.

\bibitem[Lee \& Joo(2023)Lee and Joo]{lee2023lama}
Jiye Lee and Hanbyul Joo.
\newblock Locomotion-action-manipulation: Synthesizing human-scene interactions in complex 3d environments.
\newblock In \emph{ICCV}, 2023.

\bibitem[Li et~al.(2024)Li, Zhang, Wong, Gokmen, Srivastava, Mart{\'\i}n-Mart{\'\i}n, Wang, Levine, Ai, Martinez, et~al.]{li2024behavior}
Chengshu Li, Ruohan Zhang, Josiah Wong, Cem Gokmen, Sanjana Srivastava, Roberto Mart{\'\i}n-Mart{\'\i}n, Chen Wang, Gabrael Levine, Wensi Ai, Benjamin Martinez, et~al.
\newblock Behavior-1k: A human-centered, embodied ai benchmark with 1,000 everyday activities and realistic simulation.
\newblock \emph{arXiv preprint arXiv:2403.09227}, 2024.

\bibitem[Ling et~al.(2020)Ling, Zinno, Cheng, and Van De~Panne]{ling2020character}
Hung~Yu Ling, Fabio Zinno, George Cheng, and Michiel Van De~Panne.
\newblock Character controllers using motion vaes.
\newblock \emph{ACM Transactions on Graphics (TOG)}, 39\penalty0 (4):\penalty0 40--1, 2020.

\bibitem[Loper et~al.(2015)Loper, Mahmood, Romero, Pons-Moll, and Black]{SMPL:2015}
Matthew Loper, Naureen Mahmood, Javier Romero, Gerard Pons-Moll, and Michael~J. Black.
\newblock {SMPL}: A skinned multi-person linear model.
\newblock \emph{SIGGRAPH Asia}, 2015.

\bibitem[Luiten et~al.(2024)Luiten, Kopanas, Leibe, and Ramanan]{luiten2023dynamic}
Jonathon Luiten, Georgios Kopanas, Bastian Leibe, and Deva Ramanan.
\newblock {Dynamic 3D Gaussians: Tracking by Persistent Dynamic View Synthesis}.
\newblock \emph{3DV}, 2024.

\bibitem[Luo et~al.(2022)Luo, Xu, Jiang, Zhou, Qiu, Zhang, Yang, Xu, and Yu]{luo2022artemis}
Haimin Luo, Teng Xu, Yuheng Jiang, Chenglin Zhou, Qiwei Qiu, Yingliang Zhang, Wei Yang, Lan Xu, and Jingyi Yu.
\newblock Artemis: articulated neural pets with appearance and motion synthesis.
\newblock \emph{ACM Transactions on Graphics (TOG)}, 41\penalty0 (4):\penalty0 1--19, 2022.

\bibitem[Ma et~al.(2017)Ma, Huang, Lee, and Kitani]{ma2017forecasting}
Wei-Chiu Ma, De-An Huang, Namhoon Lee, and Kris~M Kitani.
\newblock Forecasting interactive dynamics of pedestrians with fictitious play.
\newblock In \emph{CVPR}, pp.\  774--782, 2017.

\bibitem[Magnenat et~al.(1988)Magnenat, Laperri{\`e}re, and Thalmann]{magnenat1988joint}
Thalmann Magnenat, Richard Laperri{\`e}re, and Daniel Thalmann.
\newblock Joint-dependent local deformations for hand animation and object grasping.
\newblock In \emph{Proceedings of Graphics Interface'88}, pp.\  26--33. Canadian Inf. Process. Soc, 1988.

\bibitem[Mahmood et~al.(2019)Mahmood, Ghorbani, Troje, Pons-Moll, and Black]{mahmood2019amass}
Naureen Mahmood, Nima Ghorbani, Nikolaus~F Troje, Gerard Pons-Moll, and Michael~J Black.
\newblock Amass: Archive of motion capture as surface shapes.
\newblock In \emph{ICCV}, pp.\  5442--5451, 2019.

\bibitem[Mangalam et~al.(2021)Mangalam, An, Girase, and Malik]{mangalam2021goals}
Karttikeya Mangalam, Yang An, Harshayu Girase, and Jitendra Malik.
\newblock From goals, waypoints \& paths to long term human trajectory forecasting.
\newblock In \emph{ICCV}, pp.\  15233--15242, 2021.

\bibitem[Menapace et~al.(2024)Menapace, Siarohin, Lathuili{\`e}re, Achlioptas, Golyanik, Tulyakov, and Ricci]{menapace2024promptable}
Willi Menapace, Aliaksandr Siarohin, St{\'e}phane Lathuili{\`e}re, Panos Achlioptas, Vladislav Golyanik, Sergey Tulyakov, and Elisa Ricci.
\newblock Promptable game models: Text-guided game simulation via masked diffusion models.
\newblock \emph{ACM Transactions on Graphics}, 43\penalty0 (2):\penalty0 1--16, 2024.

\bibitem[Mildenhall et~al.(2020)Mildenhall, Srinivasan, Tancik, Barron, Ramamoorthi, and Ng]{mildenhall2020nerf}
Ben Mildenhall, Pratul~P Srinivasan, Matthew Tancik, Jonathan~T Barron, Ravi Ramamoorthi, and Ren Ng.
\newblock Nerf: Representing scenes as neural radiance fields for view synthesis.
\newblock In \emph{ECCV}, 2020.

\bibitem[Niemeyer \& Geiger(2021)Niemeyer and Geiger]{niemeyer2021giraffe}
Michael Niemeyer and Andreas Geiger.
\newblock Giraffe: Representing scenes as compositional generative neural feature fields.
\newblock In \emph{CVPR}, pp.\  11453--11464, 2021.

\bibitem[Oquab et~al.(2023)Oquab, Darcet, Moutakanni, Vo, Szafraniec, Khalidov, Fernandez, Haziza, Massa, El-Nouby, Howes, Huang, Xu, Sharma, Li, Galuba, Rabbat, Assran, Ballas, Synnaeve, Misra, Jegou, Mairal, Labatut, Joulin, and Bojanowski]{oquab2023dinov2}
Maxime Oquab, Timothée Darcet, Theo Moutakanni, Huy~V. Vo, Marc Szafraniec, Vasil Khalidov, Pierre Fernandez, Daniel Haziza, Francisco Massa, Alaaeldin El-Nouby, Russell Howes, Po-Yao Huang, Hu~Xu, Vasu Sharma, Shang-Wen Li, Wojciech Galuba, Mike Rabbat, Mido Assran, Nicolas Ballas, Gabriel Synnaeve, Ishan Misra, Herve Jegou, Julien Mairal, Patrick Labatut, Armand Joulin, and Piotr Bojanowski.
\newblock Dinov2: Learning robust visual features without supervision, 2023.

\bibitem[Park et~al.(2023)Park, O'Brien, Cai, Morris, Liang, and Bernstein]{park2023generative}
Joon~Sung Park, Joseph O'Brien, Carrie~Jun Cai, Meredith~Ringel Morris, Percy Liang, and Michael~S Bernstein.
\newblock Generative agents: Interactive simulacra of human behavior.
\newblock In \emph{Proceedings of the 36th Annual ACM Symposium on User Interface Software and Technology}, pp.\  1--22, 2023.

\bibitem[Park et~al.(2021)Park, Sinha, Barron, Bouaziz, Goldman, Seitz, and Martin-Brualla]{park2021nerfies}
Keunhong Park, Utkarsh Sinha, Jonathan~T. Barron, Sofien Bouaziz, Dan~B Goldman, Steven~M. Seitz, and Ricardo Martin-Brualla.
\newblock Nerfies: Deformable neural radiance fields.
\newblock In \emph{ICCV}, 2021.

\bibitem[Pavlakos et~al.(2022)Pavlakos, Weber, Tancik, and Kanazawa]{pavlakos2022one}
Georgios Pavlakos, Ethan Weber, Matthew Tancik, and Angjoo Kanazawa.
\newblock The one where they reconstructed 3d humans and environments in tv shows.
\newblock In \emph{ECCV}, pp.\  732--749. Springer, 2022.

\bibitem[Pi et~al.(2023)Pi, Peng, Yang, Zhou, and Bao]{pi2023hierarchical}
Huaijin Pi, Sida Peng, Minghui Yang, Xiaowei Zhou, and Hujun Bao.
\newblock Hierarchical generation of human-object interactions with diffusion probabilistic models.
\newblock In \emph{ICCV}, pp.\  15061--15073, 2023.

\bibitem[Puig et~al.(2023)Puig, Undersander, Szot, Cote, Yang, Partsey, Desai, Clegg, Hlavac, Min, et~al.]{puig2023habitat}
Xavier Puig, Eric Undersander, Andrew Szot, Mikael~Dallaire Cote, Tsung-Yen Yang, Ruslan Partsey, Ruta Desai, Alexander Clegg, Michal Hlavac, So~Yeon Min, et~al.
\newblock Habitat 3.0: A co-habitat for humans, avatars, and robots.
\newblock In \emph{ICLR}, 2023.

\bibitem[Rempe et~al.(2023)Rempe, Luo, Bin~Peng, Yuan, Kitani, Kreis, Fidler, and Litany]{rempe2023trace}
Davis Rempe, Zhengyi Luo, Xue Bin~Peng, Ye~Yuan, Kris Kitani, Karsten Kreis, Sanja Fidler, and Or~Litany.
\newblock Trace and pace: Controllable pedestrian animation via guided trajectory diffusion.
\newblock In \emph{CVPR}, pp.\  13756--13766, 2023.

\bibitem[Rhinehart \& Kitani(2016)Rhinehart and Kitani]{rhinehart2016learning}
Nicholas Rhinehart and Kris~M Kitani.
\newblock Learning action maps of large environments via first-person vision.
\newblock In \emph{CVPR}, pp.\  580--588, 2016.

\bibitem[Rhinehart et~al.(2019)Rhinehart, McAllister, Kitani, and Levine]{rhinehart2019precog}
Nicholas Rhinehart, Rowan McAllister, Kris Kitani, and Sergey Levine.
\newblock Precog: Prediction conditioned on goals in visual multi-agent settings.
\newblock In \emph{ICCV}, pp.\  2821--2830, 2019.

\bibitem[Saito et~al.(2021)Saito, Yang, Ma, and Black]{saito2021scanimate}
Shunsuke Saito, Jinlong Yang, Qianli Ma, and Michael~J Black.
\newblock Scanimate: Weakly supervised learning of skinned clothed avatar networks.
\newblock In \emph{CVPR}, pp.\  2886--2897, 2021.

\bibitem[Salzmann et~al.(2020)Salzmann, Ivanovic, Chakravarty, and Pavone]{salzmann2020trajectron++}
Tim Salzmann, Boris Ivanovic, Punarjay Chakravarty, and Marco Pavone.
\newblock Trajectron++: Dynamically-feasible trajectory forecasting with heterogeneous data.
\newblock In \emph{ECCV}, pp.\  683--700. Springer, 2020.

\bibitem[Sarlin et~al.(2019)Sarlin, Cadena, Siegwart, and Dymczyk]{sarlin2019coarse}
Paul-Edouard Sarlin, Cesar Cadena, Roland Siegwart, and Marcin Dymczyk.
\newblock From coarse to fine: Robust hierarchical localization at large scale.
\newblock In \emph{CVPR}, pp.\  12716--12725, 2019.

\bibitem[Seff et~al.(2023)Seff, Cera, Chen, Ng, Zhou, Nayakanti, Refaat, Al-Rfou, and Sapp]{seff2023motionlm}
Ari Seff, Brian Cera, Dian Chen, Mason Ng, Aurick Zhou, Nigamaa Nayakanti, Khaled~S Refaat, Rami Al-Rfou, and Benjamin Sapp.
\newblock Motionlm: Multi-agent motion forecasting as language modeling.
\newblock In \emph{ICCV}, pp.\  8579--8590, 2023.

\bibitem[Shafir et~al.(2023)Shafir, Tevet, Kapon, and Bermano]{shafir2023human}
Yonatan Shafir, Guy Tevet, Roy Kapon, and Amit~H Bermano.
\newblock Human motion diffusion as a generative prior.
\newblock \emph{arXiv preprint arXiv:2303.01418}, 2023.

\bibitem[Song et~al.(2023)Song, Yang, Deng, Zhu, and Ramanan]{song2023totalrecon}
Chonghyuk Song, Gengshan Yang, Kangle Deng, Jun-Yan Zhu, and Deva Ramanan.
\newblock Total-recon: Deformable scene reconstruction for embodied view synthesis.
\newblock In \emph{ICCV}, 2023.

\bibitem[Song et~al.(2020)Song, Sohl-Dickstein, Kingma, Kumar, Ermon, and Poole]{song2020score}
Yang Song, Jascha Sohl-Dickstein, Diederik~P Kingma, Abhishek Kumar, Stefano Ermon, and Ben Poole.
\newblock Score-based generative modeling through stochastic differential equations.
\newblock \emph{arXiv preprint arXiv:2011.13456}, 2020.

\bibitem[Srivastava et~al.(2022)Srivastava, Li, Lingelbach, Mart{\'\i}n-Mart{\'\i}n, Xia, Vainio, Lian, Gokmen, Buch, Liu, et~al.]{srivastava2022behavior}
Sanjana Srivastava, Chengshu Li, Michael Lingelbach, Roberto Mart{\'\i}n-Mart{\'\i}n, Fei Xia, Kent~Elliott Vainio, Zheng Lian, Cem Gokmen, Shyamal Buch, Karen Liu, et~al.
\newblock Behavior: Benchmark for everyday household activities in virtual, interactive, and ecological environments.
\newblock In \emph{CoRL}, pp.\  477--490, 2022.

\bibitem[Starke et~al.(2022)Starke, Mason, and Komura]{starke2022deepphase}
Sebastian Starke, Ian Mason, and Taku Komura.
\newblock Deepphase: Periodic autoencoders for learning motion phase manifolds.
\newblock \emph{ACM Transactions on Graphics (TOG)}, 41\penalty0 (4):\penalty0 1--13, 2022.

\bibitem[Sun et~al.(2023)Sun, Hao, Huang, Savarese, Schindler, Pollefeys, and Armeni]{sun2023nothing}
Tao Sun, Yan Hao, Shengyu Huang, Silvio Savarese, Konrad Schindler, Marc Pollefeys, and Iro Armeni.
\newblock Nothing stands still: A spatiotemporal benchmark on 3d point cloud registration under large geometric and temporal change.
\newblock \emph{arXiv preprint arXiv:2311.09346}, 2023.

\bibitem[Szeliski \& Kang(1997)Szeliski and Kang]{szeliski1997shape}
Richard Szeliski and Sing~Bing Kang.
\newblock Shape ambiguities in structure from motion.
\newblock \emph{TPAMI}, 19\penalty0 (5):\penalty0 506--512, 1997.

\bibitem[Tevet et~al.(2022)Tevet, Raab, Gordon, Shafir, Cohen-Or, and Bermano]{tevet2022human}
Guy Tevet, Sigal Raab, Brian Gordon, Yonatan Shafir, Daniel Cohen-Or, and Amit~H Bermano.
\newblock Human motion diffusion model.
\newblock \emph{arXiv preprint arXiv:2209.14916}, 2022.

\bibitem[Van Den~Berg et~al.(2011)Van Den~Berg, Guy, Lin, and Manocha]{van2011reciprocal}
Jur Van Den~Berg, Stephen~J Guy, Ming Lin, and Dinesh Manocha.
\newblock Reciprocal n-body collision avoidance.
\newblock In \emph{Robotics Research: The 14th International Symposium ISRR}, pp.\  3--19. Springer, 2011.

\bibitem[Wang et~al.(2023)Wang, Leroy, Cabon, Chidlovskii, and Revaud]{wang2023dust3r}
Shuzhe Wang, Vincent Leroy, Yohann Cabon, Boris Chidlovskii, and Jerome Revaud.
\newblock Dust3r: Geometric 3d vision made easy.
\newblock \emph{arXiv preprint arXiv:2312.14132}, 2023.

\bibitem[Wu et~al.(2023)Wu, Mildenhall, Henzler, Park, Gao, Watson, Srinivasan, Verbin, Barron, Poole, et~al.]{wu2023reconfusion}
Rundi Wu, Ben Mildenhall, Philipp Henzler, Keunhong Park, Ruiqi Gao, Daniel Watson, Pratul~P Srinivasan, Dor Verbin, Jonathan~T Barron, Ben Poole, et~al.
\newblock Reconfusion: 3d reconstruction with diffusion priors.
\newblock \emph{arXiv preprint arXiv:2312.02981}, 2023.

\bibitem[Wu et~al.(2021)Wu, Jakab, Rupprecht, and Vedaldi]{wu2021dove}
Shangzhe Wu, Tomas Jakab, Christian Rupprecht, and Andrea Vedaldi.
\newblock Dove: Learning deformable 3d objects by watching videos.
\newblock \emph{arXiv preprint arXiv:2107.10844}, 2021.

\bibitem[Xie et~al.(2023)Xie, Jampani, Zhong, Sun, and Jiang]{xie2023omnicontrol}
Yiming Xie, Varun Jampani, Lei Zhong, Deqing Sun, and Huaizu Jiang.
\newblock Omnicontrol: Control any joint at any time for human motion generation.
\newblock \emph{arXiv preprint arXiv:2310.08580}, 2023.

\bibitem[Yang \& Ramanan(2019)Yang and Ramanan]{yang2019volumetric}
Gengshan Yang and Deva Ramanan.
\newblock Volumetric correspondence networks for optical flow.
\newblock In \emph{NeurIPS}, 2019.

\bibitem[Yang et~al.(2022)Yang, Vo, Natalia, Ramanan, Vedaldi, and Joo]{yang2022banmo}
Gengshan Yang, Minh Vo, Neverova Natalia, Deva Ramanan, Andrea Vedaldi, and Hanbyul Joo.
\newblock Banmo: Building animatable 3d neural models from many casual videos.
\newblock In \emph{CVPR}, 2022.

\bibitem[Yang et~al.(2023{\natexlab{a}})Yang, Tan, Lyons, Peri, and Ramanan]{Yang_Lab4d_-_A_2023}
Gengshan Yang, Jeff Tan, Alex Lyons, Neehar Peri, and Deva Ramanan.
\newblock {Lab4d - A framework for in-the-wild 4D reconstruction from monocular videos}, June 2023{\natexlab{a}}.
\newblock URL \url{https://github.com/lab4d-org/lab4d}.

\bibitem[Yang et~al.(2023{\natexlab{b}})Yang, Gao, Li, Gao, Wang, and Zheng]{yang2023track}
Jinyu Yang, Mingqi Gao, Zhe Li, Shang Gao, Fangjing Wang, and Feng Zheng.
\newblock Track anything: Segment anything meets videos, 2023{\natexlab{b}}.

\bibitem[Ye et~al.(2023)Ye, Pavlakos, Malik, and Kanazawa]{ye2023decoupling}
Vickie Ye, Georgios Pavlakos, Jitendra Malik, and Angjoo Kanazawa.
\newblock Decoupling human and camera motion from videos in the wild.
\newblock In \emph{CVPR}, pp.\  21222--21232, 2023.

\bibitem[Yuan et~al.(2022)Yuan, Iqbal, Molchanov, Kitani, and Kautz]{yuan2022glamr}
Ye~Yuan, Umar Iqbal, Pavlo Molchanov, Kris Kitani, and Jan Kautz.
\newblock Glamr: Global occlusion-aware human mesh recovery with dynamic cameras.
\newblock In \emph{CVPR}, pp.\  11038--11049, 2022.

\bibitem[Yuan et~al.(2023)Yuan, Song, Iqbal, Vahdat, and Kautz]{yuan2023physdiff}
Ye~Yuan, Jiaming Song, Umar Iqbal, Arash Vahdat, and Jan Kautz.
\newblock Physdiff: Physics-guided human motion diffusion model.
\newblock In \emph{ICCV}, pp.\  16010--16021, 2023.

\bibitem[Zhang et~al.(2023{\natexlab{a}})Zhang, Yuan, Makoviychuk, Guo, Fidler, Peng, and Fatahalian]{zhang2023vid2player3d}
Haotian Zhang, Ye~Yuan, Viktor Makoviychuk, Yunrong Guo, Sanja Fidler, Xue~Bin Peng, and Kayvon Fatahalian.
\newblock Learning physically simulated tennis skills from broadcast videos.
\newblock \emph{ACM Transactions on Graphics (TOG)}, 42\penalty0 (4):\penalty0 1--14, 2023{\natexlab{a}}.

\bibitem[Zhang et~al.(2018)Zhang, Starke, Komura, and Saito]{zhang2018mode}
He~Zhang, Sebastian Starke, Taku Komura, and Jun Saito.
\newblock Mode-adaptive neural networks for quadruped motion control.
\newblock \emph{ACM Transactions on Graphics (TOG)}, 37\penalty0 (4):\penalty0 1--11, 2018.

\bibitem[Zhang et~al.(2023{\natexlab{b}})Zhang, Rao, and Agrawala]{zhang2023adding}
Lvmin Zhang, Anyi Rao, and Maneesh Agrawala.
\newblock Adding conditional control to text-to-image diffusion models.
\newblock In \emph{ICCV}, 2023{\natexlab{b}}.

\bibitem[Zhao et~al.(2023)Zhao, Zhang, Wang, Beeler, and Tang]{zhao2023synthesizing}
Kaifeng Zhao, Yan Zhang, Shaofei Wang, Thabo Beeler, and Siyu Tang.
\newblock Synthesizing diverse human motions in 3d indoor scenes.
\newblock \emph{arXiv preprint arXiv:2305.12411}, 2023.

\bibitem[Zhong et~al.(2023)Zhong, Rempe, Xu, Chen, Veer, Che, Ray, and Pavone]{zhong2023guided}
Ziyuan Zhong, Davis Rempe, Danfei Xu, Yuxiao Chen, Sushant Veer, Tong Che, Baishakhi Ray, and Marco Pavone.
\newblock Guided conditional diffusion for controllable traffic simulation.
\newblock In \emph{ICRA}, pp.\  3560--3566. IEEE, 2023.

\bibitem[Ziebart et~al.(2008)Ziebart, Maas, Bagnell, Dey, et~al.]{ziebart2008maximum}
Brian~D Ziebart, Andrew~L Maas, J~Andrew Bagnell, Anind~K Dey, et~al.
\newblock Maximum entropy inverse reinforcement learning.
\newblock In \emph{AAAI}, volume~8, pp.\  1433--1438. Chicago, IL, USA, 2008.

\bibitem[Ziebart et~al.(2009)Ziebart, Ratliff, Gallagher, Mertz, Peterson, Bagnell, Hebert, Dey, and Srinivasa]{ziebart2009planning}
Brian~D Ziebart, Nathan Ratliff, Garratt Gallagher, Christoph Mertz, Kevin Peterson, J~Andrew Bagnell, Martial Hebert, Anind~K Dey, and Siddhartha Srinivasa.
\newblock Planning-based prediction for pedestrians.
\newblock In \emph{IROS}, pp.\  3931--3936. IEEE, 2009.

\end{thebibliography}
\bibliographystyle{iclr2025_conference}

\clearpage
\newpage
\appendix
\section{Appendix}

\subsection{Refinement with 3D Gaussians}
On the project webpage, we show results of rendering the generated interactions with 3D Gaussians~\citep{kerbl20233d}. We explain how this is achieved below.

\noindent{\bf Representation.} Neural implicit representations are suitable for coarse optimization, but can be difficult to converge and slow to render. Therefore, we introduce a refinement procedure that replaces the canonical shape model ${\bf T}$ with 3D Gaussians while keeping the motion model $\mathcal{D}$ as is. 

We use 20k Gaussians for the agent and 200k Gaussians for the scene, each parameterized by 14 values, including its opacity, RGB color, center location, orientation, and axis-aligned scales. To render an image, we warp the agent Gaussians forward from canonical space to time $t$ (Eq.~\ref{eq:lbs}), compose them with the scene Gaussians, and call the differentiable Gaussian rasterizer.

\noindent{\bf Optimization.} Gaussians are initialized with points on a mesh extracted from Eq.~(\ref{eq:density_env}-\ref{eq:density_agent}) and assigned isotropic scales. To initialize the color of each Gaussian, we query the canonical color MLP at its center. We update both the canonical 3D Gaussian parameters and the motion fields by minimizing
\begin{equation}\label{eq:loss-gs}    
    \min_{{\bf T}, \mathcal{D}} \sum_{t}\|I_t - \mathcal{R}_{I}(t; {\bf T}, \mathcal{D})\|_2^2 + L_{reg}({\bf T}, \mathcal{D}),
\end{equation}
where the regularization term includes a flow loss, a depth loss and a silhouette loss on the agent.

\subsection{Details on Model and Data}

\noindent{\bf Table of Notation.} A table of notation used in the paper can be found in Tab.~\ref{tab:parameters}.

\noindent{\bf Summary of I/O.} A summary of inputs and outputs of the method is shown in Tab.~\ref{tab:io} 

\begin{table*}
\caption{Table of Notation.}
\centering
\makebox[\textwidth]{
\begin{tabular}{lll}
\toprule
 Symbol  & Description \\ 
 \midrule
 \multicolumn{2}{c}{\bf Global Notations}\\
 $B$ & The number of bones of an agent. By defatult $B=25$.\\
 $M$ & The number of videos. \\
 $N_i$ & The number of image frames extracted from video $i$. \\
 ${\bf I}_i$ & The sequence of color images $\{I_1, \dots, I_{N_i}\}$ extracted from video $i$. \\
 $\boldsymbol{\psi}_i$        & The sequence of DINOv2 feature images $\{\boldsymbol{\psi}_1, \dots,\boldsymbol{\psi}_{N_i} \}$ extracted from video $i$. \\  
 $T_i$ & The length of video $i$. \\
 $T^*$ & The time horizon of behavior diffusion. By default $T^*=5.6$s.\\
 $T'$ & The time horizon of past conditioning. By default $T'=0.8$s\\
 ${\bf Z}\in \mathbb{R}^3$ &  Goal of the agent, defined as the location at the end of $T^*$.  \\
 ${\bf P}\in \mathbb{R}^{3 \times T^*}$ & Path of the agent, defined as the root body trajectory over $T^*$. \\
 ${\bf G}\in \mathbb{R}^{6B\times T^*}$ & Pose of the agent, defined as the 6DoF rigid motion of bones over $T^*$.\\
 $\omega_{s}\in\mathbb{R}^{64}$& Scene code, representing the scene perceived by the agent.\\
 $\omega_{o}\in\mathbb{R}^{64}$& Observer code, representing the observer perceived by the agent.\\
 $\omega_{p}\in\mathbb{R}^{64}$& Past code, representing the history of events happened to the agent. \\
 \midrule
 \multicolumn{2}{c}{\bf Learnable Parameters of 4D Reconstruction}\\
  ${\bf T}$           & Canonical NeRFs, including a scene MLP and an agent MLP.\\
  $\boldsymbol{\beta}_i\in\mathbb{R}^{128}$ & Per-video code that allows NeRFs to represent variations across videos.\\
  $\mathcal{D}$ & Time-varying parameters,  including $\{{\boldsymbol{\xi}}, {\bf G}, {\bf W}\}$.\\
  ${\boldsymbol{\xi}}_t\in SE(3)$ & The camera pose that transforms the scene to the camera coordinates at t. \\
  ${\bf G}^0_t \in SE(3)$ & The camera pose that transforms the canonical agent to the camera coordinates at t. \\
  ${\bf G}^b_t \in SE(3)$ & The transformation that moves bone b from its rest state to time t state. \\
 ${\bf W}\in\mathbb{R}^B$                  & Skinning weights of a point, defined as the probability of belonging to bones.\\
 $f_{\theta}$ & PoseNet that takes a DINOv2 feature image as input and produces camera pose.\\
  \midrule
 \multicolumn{2}{c}{\bf Learnable Parameters of Behavior Generation}\\
$\mathrm{MLP}_{\theta_{\bf Z}}$ & Goal MLP that represent the score function of goal distributions. \\
$\mathrm{ControlUNet}_{\theta_{\bf P}}$ & Path UNet that represents the score function of path distributions.\\
$\mathrm{ControlUNet}_{\theta_{\bf G}}$ & Pose UNet that represents the score function of pose distributions.\\
$\mathrm{ResNet3D}_{\theta_{\boldsymbol{\psi}}}$ & Scene perception network that produces $\omega_s$ from 3D feature grids $\boldsymbol{\psi}$. \\
$\mathrm{MLP}_{\theta_{\bf o}}$ & Observer MLP that produces $\omega_{o}$ from observer's past trajectory in $T'$.\\
$\mathrm{MLP}_{\theta_{\bf p}}$ & Past MLP that produces $\omega_{p}$ from agent's past trajectory in $T'$.\\
 \bottomrule
\label{tab:parameters}
\end{tabular}
}
\end{table*}

\begin{table*}
\caption{Summary of inputs and outputs at different stages of the method.}
\centering
\makebox[\textwidth]{
\begin{tabular}{lll}
\toprule
 Stage  & Description \\ 
 \midrule
 Overall & Input: A walk-through video of the scene and videos with agent interactions.\\
 & Output: An interactive behavior generator of the agent. \\
 \midrule
Localizer Training & Input: 3D reconstruction of the environment and the agent.\\
& Output: Neural localizer  $f_\theta$. \\
\midrule
Neural Localization & Input: Neural localizer  $f_\theta$ and the agent interaction videos.\\
& Output: Camera poses for each video frame.\\
 \midrule
4D Reconstruction & Input: A collection of videos and their corresponding camera poses.\\
& Output: Scene feature volume $\boldsymbol{\Psi}$, motion of the agent {\bf G} and observer ${\boldsymbol \xi}$.\\
 \midrule
Behavior Learning & Input: Scene feature volume $\boldsymbol{\Psi}$, motion of the agent {\bf G} and observer ${\boldsymbol \xi}$.\\
& Output: An interactive behavior generator of the agent. \\
 \bottomrule
\label{tab:io}
\end{tabular}
}
\end{table*}

\noindent{\bf Data Collection.} We collect RGBD videos using an iPhone, similar to TotalRecon~\citep{song2023totalrecon}. To train the neural localizer, we use Polycam to take the walkthrough video and extract a textured mesh. For behavior capture, we use Record3D App to record videos and extract color images and depth images.

\noindent{\bf Diffusion Model Architecture.} The score function of the goal is implemented as 6-layer MLP with hidden size 128. The the score functions of the paths and body motions are implemented as 1D UNets taken from GMD~\citep{karunratanakul2023guided}. The sampling frequency is set to be $0.1$s, resulting a sequence length of $56$. The environment encoder is implemented as a 6-layer 3D ConvNet with kernel size 3 and channel dimension 128. The observer encoder and history encoder are implemented as a 3-layer MLP with hidden size 128. 

\begin{figure}[!h]
    \centering
    \includegraphics[width=\linewidth,trim={3.5cm 8.5cm 3.5cm 8.5cm},clip]{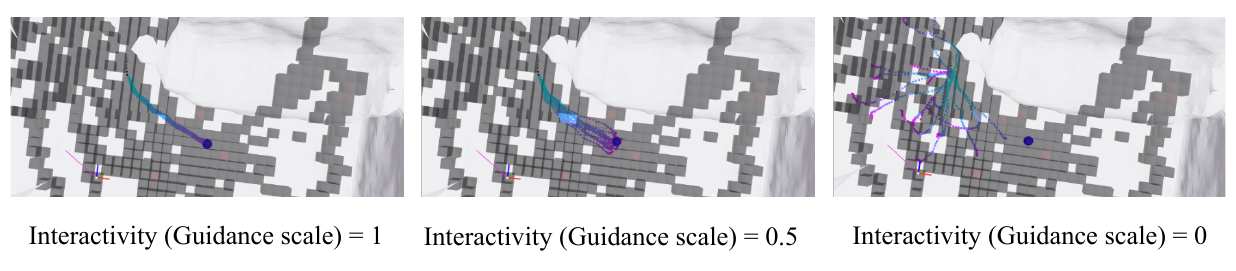}
    \caption{{\bf Interactivity of the agent}. By changing the classifier-free guidance scale $s$, we can find a trade-off between interactive behavior and unconditional behavior. We demonstrate the control over interactivity by goal-conditioned path generation (bird's-eye-view, blue sphere$\rightarrow$goal; gradient color$\rightarrow$generated path). With a higher classifier-free guidance scale $s$, the model is controlled more by the conditional generator, and therefore exhibits higher interactivity. $s=0$ corresponds to fully unconditional generation.}
    \label{fig:affinity}
\end{figure}

\noindent{\bf Diffusion Model Training and Testing.} We use a linear noise schedule at training time and $50$ denoising steps. We train all the diffusion models (goal, path and pose) with classifier-free guidance~\citep{ho2022classifier, tevet2022human} that randomly sets conditioning signals to zeros ${\bf Z} = \varnothing$ randomly. This allows us to control the trade-off between interactive behavior and unconditional behavior generation, as shown in Fig.~\ref{fig:affinity}. At test time, each goal denoising step takes $2$ms and each path/body denoising step takes $9$ms on an A100 GPU.

\subsection{Additional Results}

\begin{figure}[h!]
\centering
\includegraphics[width= \linewidth,trim={6cm 5.8cm 6cm 5.8cm},clip]{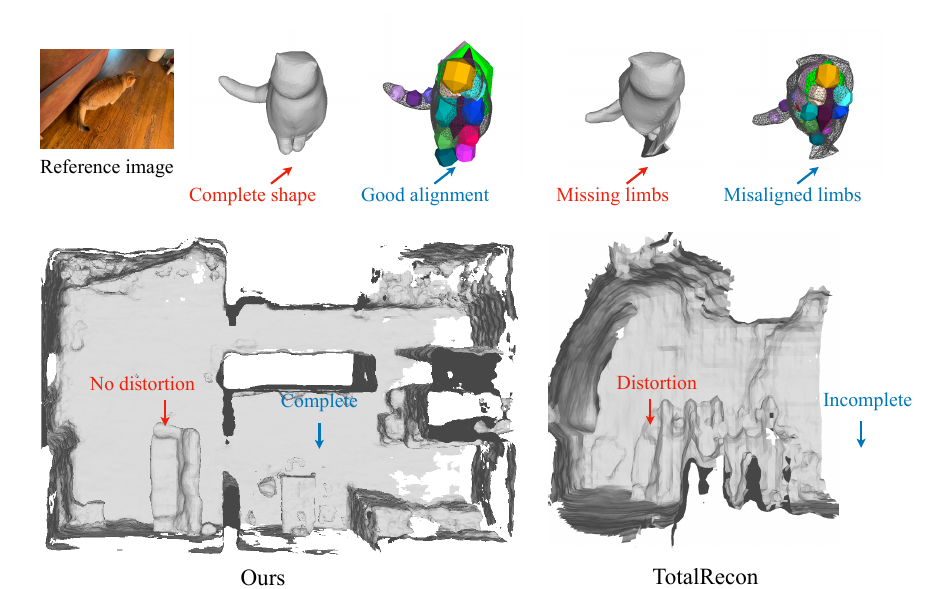}
\caption{Qualitative comparison with TotalRecon~\citep{song2023totalrecon} on 4D reconstruction. Top: reconstruction of the agent at at specific frame. Total-recon produces shapes with missing limbs and bone transformations that are misaligned with the shape, while our method produces complete shapes and good alignment. Bottom: reconstruction of the environment. TotalRecon produces distorted and incomplete geometry (due to lack of observations from a single video), while our method produces an accurate and complete environment reconstruction.}
\label{fig:vstotalrecon}
\end{figure}

\noindent{\bf Comparison to TotalRecon.}
In the main paper, we compare to TotalRecon on scene reconstruction by providing it multiple videos. Here, we include additional comparison in their the original single video setup. We find that TotalRecon fails to build a good agent model, or a complete scene model given limited observations, while our method can leverage multiple videos as inputs to build a better agent and scene model. The results are shown in Fig.~\ref{fig:vstotalrecon}.

\begin{figure*}
    \centering
    \includegraphics[width=\linewidth, trim={0 0 0 0},clip]{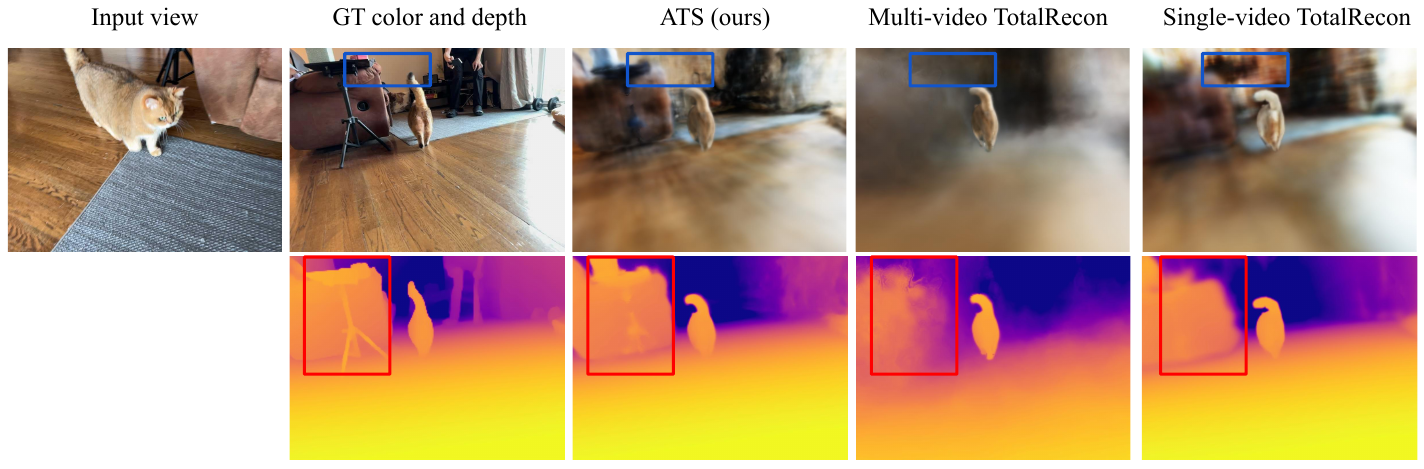}
    \caption{{\bf Qualitative comparison on 4D reconstruction (Tab.~\ref{tab:4dreconresults}).} We compare with TotalRecon on 4D reconstruction quality. We show novel views rendered with a held-out camera that looks from the opposite side. ATS is able to leverage multiple videos captured at different times to reconstruct the wall (blue box) and the tripod stand (red box) even they are not visible in the input views. Multi-video TotalRecon produces blurry RGB and depth due to bad camera registration. The original TotalRecon takes a single video as input and therefore fails to reconstruct the regions (the tripod and the wall) that are not visible in the input video. }
    \label{fig:4d-eval}
\end{figure*}

\begin{figure}[ht]
\centering
\includegraphics[width=\linewidth,trim={2cm 3.5cm 2cm 3.5cm},clip]{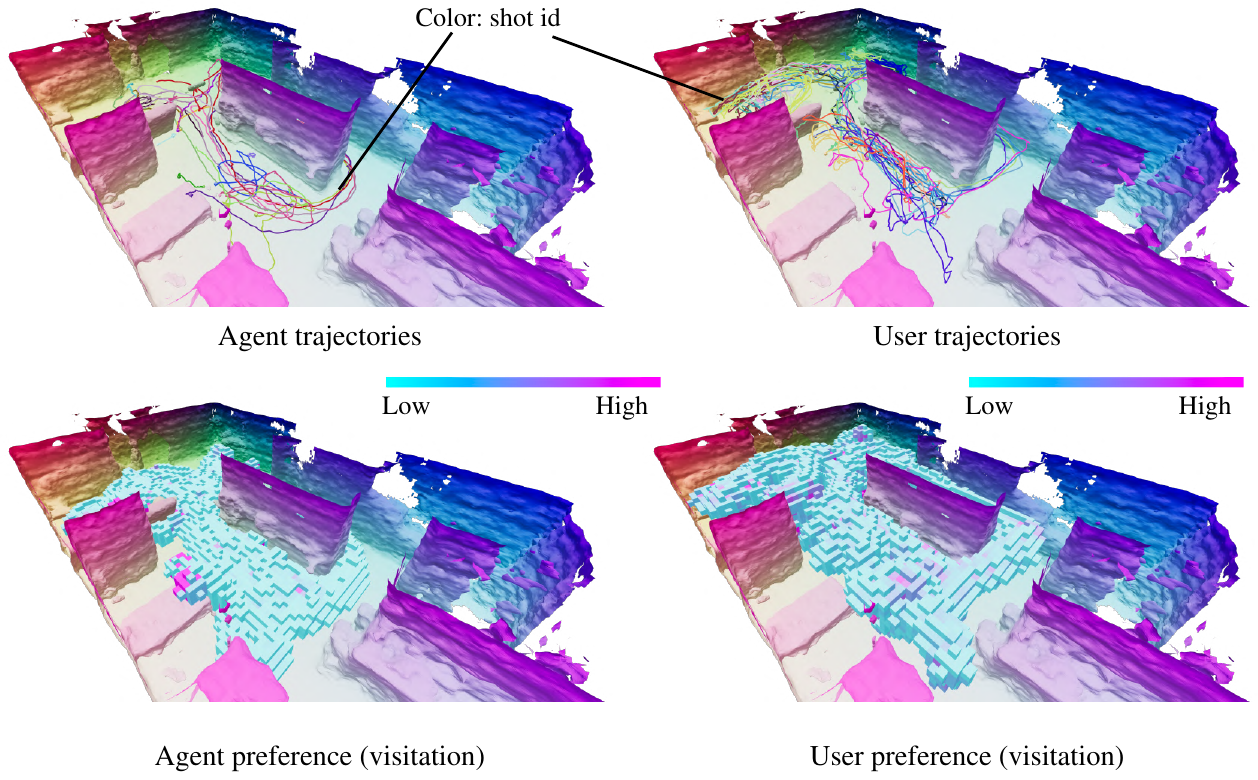}
\caption{Given the 3D trajectories of the agent and the user accumulated over time (top), one could compute their preference represented by 3D heatmaps (bottom). Note the high agent preference over table and sofa.}
\label{fig:align-recon}
\end{figure}

\noindent{\bf Visual Ablation on Scene Awareness.}
We show final camera and agent registration to the canonical scene in Fig.~\ref{fig:align-recon}. The registered 3D trajectories provides statistics of agent's and user's preference over the environment. 

\begin{figure}[t!]
\centering
\includegraphics[width=\linewidth,trim={0cm 5cm 0cm 5cm},clip]{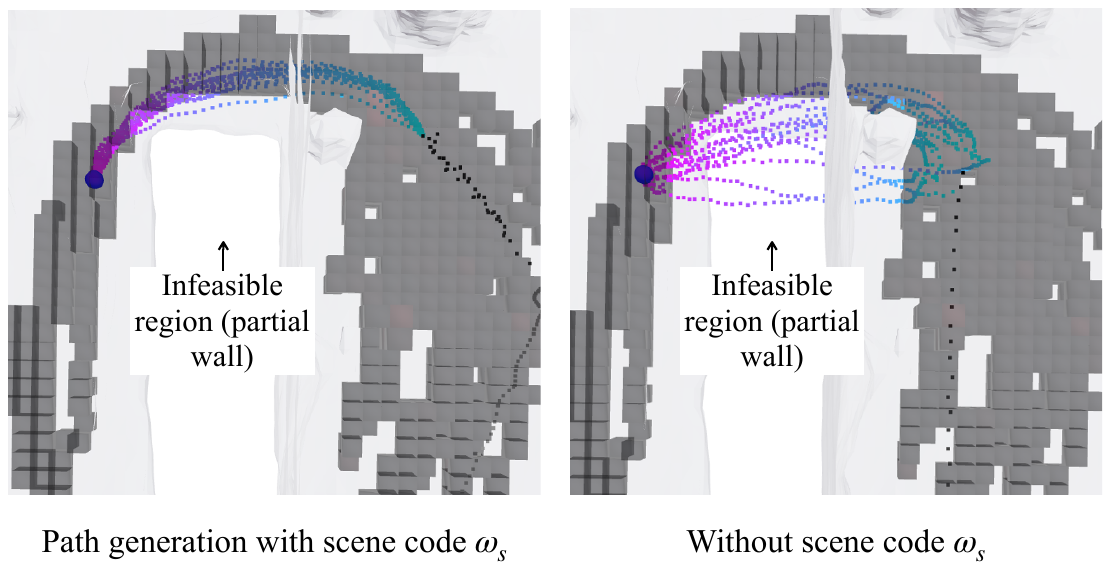}
\caption{{\bf Visual ablation on scene awareness.} We demonstrate the effect of the scene code $\omega_s$ through goal-conditioned path generation (bird's-eye-view, blue sphere$\rightarrow$goal; gradient color$\rightarrow$generated path; gray blocks$\rightarrow$locations that have been visited in the training data). Conditioned on scene, the generated path abide by the scene geometry, while removing the scene code, the generated paths go through the wall in between two empty spaces.}
\label{fig:scene-awareness}
\end{figure}

\noindent{\bf Histogram of Agent / Observer Visitation.}
We show final camera and agent registration to the canonical scene in Fig.~\ref{fig:scene-awareness}. The registered 3D trajectories provides statistics of agent's and user's preference over the environment. 
 
\begin{figure*}
    \centering
    \includegraphics[width=\linewidth, trim={0 0 0 0},clip]{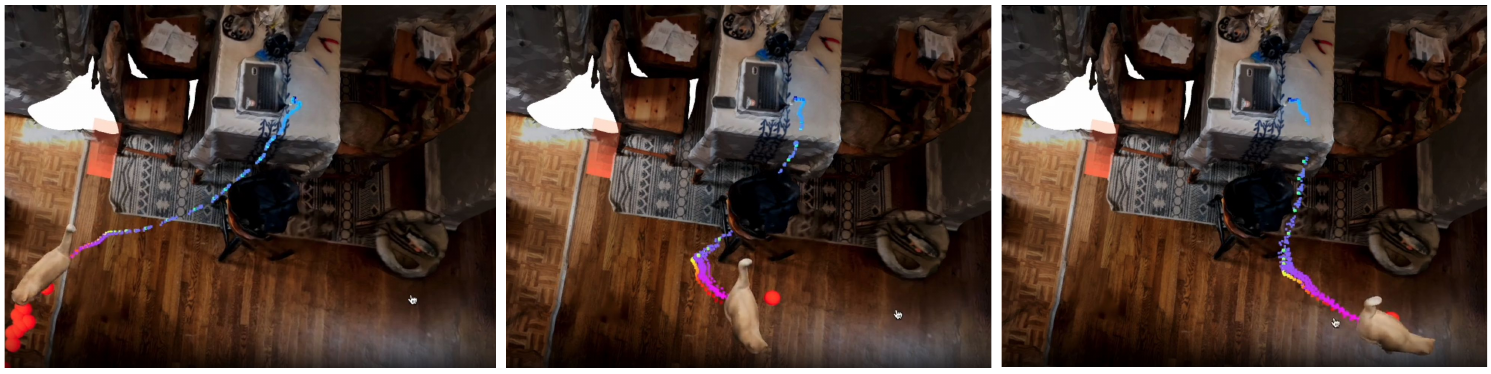}
    \caption{{\bf Generalization ability of the behavior model.} Thanks to the ego-centric encoding design (Eq.~12), a specific behavior can be learned and generalized to novel situations even it was seen once. Although there's only one data point where the cat jumps off the dining table, our method can generate diverse motion of cat jumping off the table while landing at different locations (to the left, middle, and right of the table) as shown in the visual. }
    \label{fig:generalization}
\end{figure*}

\subsection{Limitations and Future Works}

\begin{figure*}[ht]
    \centering
    \begin{minipage}{0.55\textwidth}
        \centering
        \includegraphics[width=\linewidth]{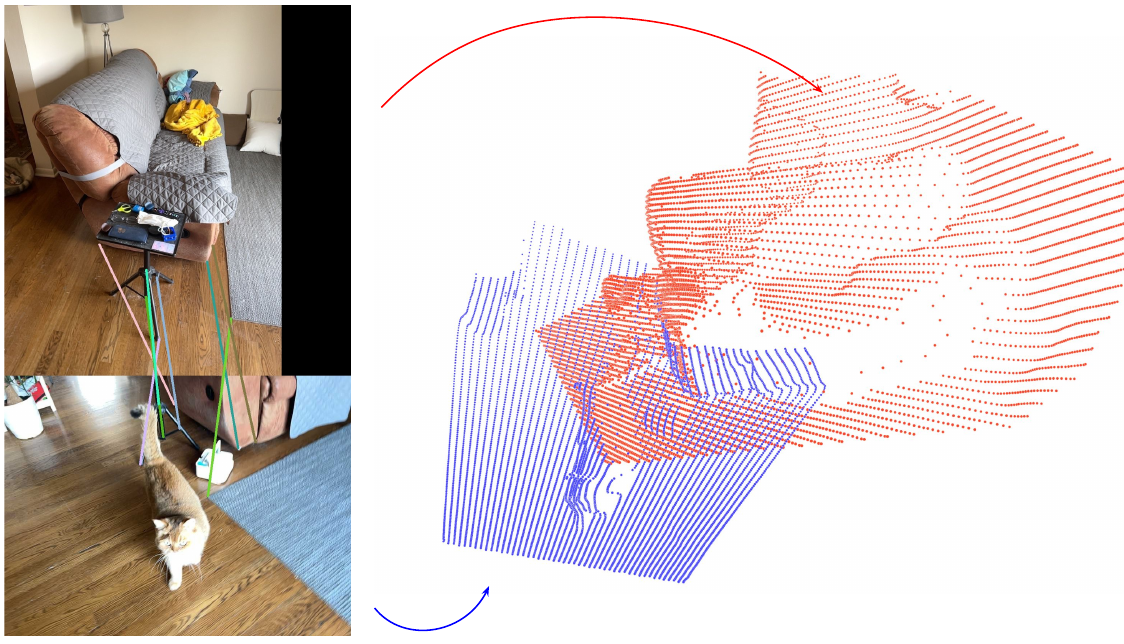}
        \caption{{\bf GT correspondence and 3D alignment.} Left: Annotated 2D correspondence between the canonical scene (top) and the input image (bottom). Right: we visualize the GT camera registration by transforming the input frame 3D points (blue, back-projected from depth) to the canonical frame (red). The points align visually.}
        \label{fig:corresp-anno}
    \end{minipage}\hfill
    \begin{minipage}{0.41\textwidth}
        \centering
        \includegraphics[width=\linewidth]{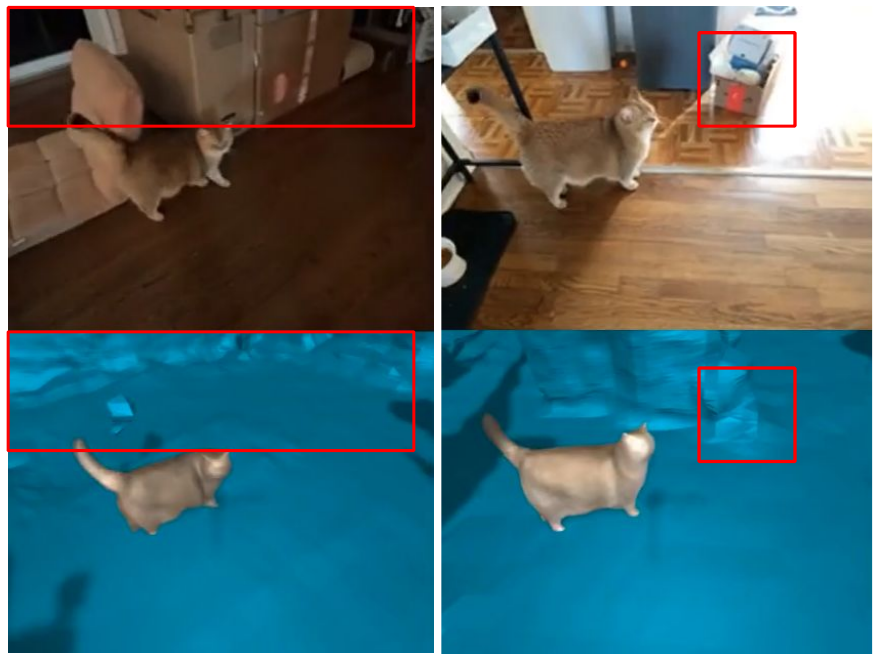}
        \caption{{\bf Robustness to layout changes.} We find our camera localization to be robust to layout changes, e.g., the cushion and the large boxes (left) and the box (right). However, it fails to \emph{reconstruct} layout changes, especially when they are only observed in a few views.}
        \label{fig:scene-failure}
    \end{minipage}
\end{figure*}

\noindent{\bf Environment Reconstruction.} To build a complete reconstruction of the environment, we register multiple videos to a shared canonical space. However, the transient structures (e.g., cushion that can be moved over time) may not be reconstructed well due to lack of observations. We notice displacement of chairs and appearance of new furniture in our capture data. Our method is robust to these in terms of camera localization (Tab.~\ref{tab:camera_registration} and Fig.~\ref{fig:scene-failure}). However, 3D reconstruction of these transient components is challenging. As shown in Fig~\ref{fig:scene-failure}, our method fails to reconstruct notable layout changes when they are only observed in a few views, e.g., the cushion and the large boxes (left) and the box (right). We leave this as future work. Leveraging generative image prior to in-paint the missing regions is a promising direction to tackle this problem~\citep{wu2023reconfusion}.

\noindent{\bf Scaling-up.} We demonstrate our approach on four types of agents with different morphology living in different environments. For the cat, we use 23 video clips over a span of a month. This isn't large-scale but we believe this is an important step to go beyond a single video.  In terms of robustness, we showed a meaningful step towards scaling up 4D reconstruction by neural initialization (Eq. 6). The major difficulty towards large-scale deployment is the cost and robustness of 4D reconstruction using test-time optimization.

\noindent{\bf Multi-agent Interactions.}  \ourmethod{} only handles interactions between the agent and the observer. Interactions with other agents in the scene are out of scope, as it requires data containing more than one agent. Solving re-identification and multi-object tracking in 4D reconstruction will enable introducing multiple agents. We leave learning multi-agent behavior from videos as future work.

\noindent{\bf Complex Scene Interactions.} Our approach treat the background as a rigid component without accounting for movable and articulated scene structures, such as doors and drawers. To reconstruct complex interactions with the environment, one approach is to extend the scene representation to be hierarchical (with a kinematic tree), such that it consists of articulated models of interactable objects. To generate plausible interactions between the agent and the scene (e.g., opening a door), one could extend the agent representation $G$ to include both the agent and the articulated objects (e.g., door).

\noindent{\bf Physical Interactions.}  Our method reconstructs and generates the kinematics of an agent, which may produce physically-implausible results (e.g., penetration with the ground and foot sliding). One promising way to deal with this problem is to add physics constraints to the reconstruction and motion generation~\citep{yuan2023physdiff}.%

\noindent{\bf Long-term Behavior.} The current \ourmethod{} model is trained with time-horizon of $T^*=6.4$ seconds. We observe that the model only learns mid-level behaviors of an agent (e.g., trying to move to a destination; staying at a location; walking around). We hope incorporating a memory module and training with longer time horizon will enable learning higher-level behaviors of an agent.

\subsection{Social Impact}

Our method is able to learn interactive behavior from videos, which could help build simulators for autonomous driving, gaming, and movie applications. It is also capable of building personalized behavior models from casually collected video data, which can benefit users who do not have access to a motion capture studio. On the negative side, the behavior generation model could be used as ``deepfake'' and poses threats to user's privacy and social security.

\clearpage
\newpage

\end{document}